\documentclass[journal=jacsat,manuscript=article]{achemso}
\usepackage[utf8]{inputenc}
\usepackage{hyperref}
\usepackage{tikz}
\usepackage{mathtools}
\usepackage{graphicx}
\usepackage{subcaption}
\usepackage{enumitem}

\usepackage{amsmath,amssymb}

\DeclareMathOperator{\EX}{\mathbb{E}}

\author{Gabriele Scalia}
\affiliation{Department of Chemical Engineering, Massachusetts Institute of Technology, Cambridge, MA, 02139, USA}
\alsoaffiliation{Department of Electronics, Information and Bioengineering, Politecnico di Milano,
20133 Milano, Italy} 
\author{Colin A. Grambow}
\affiliation[MIT]
{Department of Chemical Engineering, Massachusetts Institute of Technology, Cambridge, MA, 02139, USA}
\author{Barbara Pernici}
\affiliation[POLIMI]
{Department of Electronics, Information and Bioengineering, Politecnico di Milano,
20133 Milano, Italy}
\author{Yi-Pei Li}
\affiliation[NTU]
{Department of Chemical Engineering, National Taiwan University, Taipei 10617, Taiwan}
\email{yipeili@ntu.edu.tw} 
\author{William H. Green}
\affiliation[MIT]
{Department of Chemical Engineering, Massachusetts Institute of Technology, Cambridge, MA, 02139, USA}
\email{whgreen@mit.edu} 

\title{Evaluating Scalable Uncertainty Estimation Methods for DNN-Based Molecular Property Prediction}

\abbreviations{QSPR}

\begin{document}

\maketitle

\section{Abstract}

Advances in deep neural network (DNN) based molecular property prediction have recently led to the development of models of remarkable accuracy and generalization ability, with graph convolution neural networks (GCNNs) reporting  state-of-the-art performance for this task.
However, some challenges remain and one of the most important that needs to be fully addressed concerns uncertainty quantification. DNN performance is affected by the volume and the quality of the training samples. Therefore, establishing when and to what extent a prediction can be considered reliable is just as important as outputting accurate predictions, especially when out-of-domain molecules are targeted.
Recently, several methods to account for uncertainty in DNNs have been proposed, most of which are based on approximate Bayesian inference. Among these, only a few scale to the large datasets required in applications.
Evaluating and comparing these methods has recently attracted great interest, but results are generally fragmented and absent for molecular property prediction.
In this paper, we aim to quantitatively compare scalable techniques for uncertainty estimation in GCNNs. We introduce a set of quantitative criteria to capture different uncertainty aspects, and then use these criteria to compare MC-Dropout, deep ensembles, and bootstrapping, both theoretically in a unified framework that separates aleatoric/epistemic uncertainty and experimentally on the QM9 dataset.
Our experiments quantify the performance of the different uncertainty estimation methods and their impact on uncertainty-related error reduction.
Our findings indicate that ensembling and bootstrapping consistently outperform MC-Dropout, with different context-specific pros and cons. Our analysis also leads to a better understanding of the role of aleatoric/epistemic uncertainty and highlights the challenge posed by out-of-domain uncertainty.

\section{Introduction}

Deep Neural Network (DNN) based molecular property prediction has received new attention recently with the development of models capable of promising performance on large and heterogeneous  datasets\cite{Yang2019,wu2018moleculenet,mayr2018large}. In particular, recent progresses in \emph{graph convolution neural network}\cite{duvenaud2015convolutional} (GCNN) --- also known as \emph{message passing neural network} (MPNN) --- have led to state-of-the-art performance for property prediction across a range of public and proprietary datasets\cite{Yang2019}, demonstrating both accuracy and generalization gains. However, some limitations still hold, and uncertainty quantification has recently been highlighted as an important direction to be investigated\cite{Yang2019}.

The need for an effective uncertainty quantification is driven by both intrinsic characteristics of DNN models and by peculiar features of chemical space. In general, standard DNN models do not output confidence estimates, since regression models only output a mean, while classification outputs cannot be reliably interpreted as confidence scores\cite{gal2016uncertainty}.

DNN performance strongly depend on the volume and the quality of training data, hence the need to assess when and to what extent a prediction can be considered reliable. While this has emerged in the context of DNN in several heterogeneous applications, most of which are based on computer vision\cite{Kendall2017}, DNN for chemistry is characterized by additional challenges. First of all, chemical training  data are intrinsically biased\cite{Zhang2019}, since the chemical space has an extremely large variability and therefore a training dataset cannot represent the whole space. Moreover, chemical training data are often limited in volume and quality, directly reflected in DNN outputs. 
Additionally, doing predictions on molecules rather different to those seen during training is often the actual goal in the field, for example in drug discovery applications. This demands good generalization performance on one side, but also being able to identify the model's \emph{knowledge boundary}, i.e. assessing to what extent the model knows what it knows.

While uncertainty estimation in this domain has been investigated in the context of shallow models in the last few years\cite{proppe2017reliable}, uncertainty in DNN and GCNN models for molecular property prediction has been addressed only recently and is still limited and fragmented. 

Bayesian Neural Networks (BNNs) have long been studied as an effective and principled way to take into account model uncertainty in the predictions of a DNN\cite{neal1995bayesian}, but the intractability of exact Bayesian inference together with the limited practicality of the approaches proposed until the last few years has prevented the widespread diffusion of these solutions in applications until recently\cite{gal2016uncertainty}. 
The recent work from \citet{gal2016dropout} gave a decisive contribution to the spread of approximate BNNs in applications, proposing Monte Carlo Dropout (MC-Dropout), a practical method based on the widely used dropout regularization technique, to account for model uncertainty. Moreover, \citet{Kendall2017} proposed a framework to separate \emph{epistemic uncertainty}, which refers to uncertainty in the model predictions, from the \emph{aleatoric uncertainty}, which captures noise inherent in the data.  MC-Dropout has been used in various applications, including, very recently, molecular property prediction\cite{Zhang2019,Ryu2018}.

Other techniques to efficiently approximate BNNs have been proposed since then, highlighting how finding a good trade-off between effective approximation and scalability remains an important open challenge. 
Notably, the ensemble-based approach proposed by \citet{lakshminarayanan2017simple} constitutes a simple and scalable technique to obtain well-calibrated uncertainty estimates and has been already used in several applications across different fields (e.g., \citet{Ron18,Tomasev2019116}).
Moreover, even if originally proposed as a non-Bayesian alternative to estimate uncertainty in DNNs\cite{lakshminarayanan2017simple}, recent work highlighted how ensembling in DNN can be traced back to Bayesian inference\cite{duvenaud2016early,pearce2018uncertainty,gustafsson2019evaluating}.

In parallel to the development of methods to efficiently approximate BNNs, their evaluation, and in particular their comparative assessment, has recently attracted great interest given the challenges it poses\cite{guo2017calibration,ilg2018uncertainty,mukhoti2018evaluating,gustafsson2019evaluating}.
Indeed, we usually do not have ``ground truth uncertainties'', which prevents using traditional benchmarks. Furthermore, evaluating uncertainty involves measuring the model's unknowns and taking into account domain-specific features.
First comparative assessments have been conducted for computer vision tasks\cite{ilg2018uncertainty,mukhoti2018evaluating,gustafsson2019evaluating}. However, results are still fragmented and no comparisons have been carried out for GCNN in the chemistry domain, which poses specific challenges  such as uncertainty generalization in the chemical space.
Moreover, many metrics traditionally used to evaluate uncertain forecasts, like \emph{calibration}, have been defined in a classification setting, while their extension for \emph{regression} --- needed for scalar molecular properties --- has been discussed only recently\cite{kuleshov2018accurate,levi2019evaluating}.

Comparative analysis of different methods calls for \emph{multiple metrics} and \emph{quantitative indices}. By contrast, recent works targeting uncertainty estimation for DNN-based molecular property prediction only employ a single technique, such as confidence-error diagrams, and qualitative evaluations\cite{Ryu2018,Zhang2019}.

The goals of this work are as follows. First of all, we review existing methods for uncertainty estimation in DNN/GCNN, focusing on scalable techniques that can be used in applications. We contextualize them in a unique framework to estimate aleatoric and epistemic uncertainty, also in light of their recent interpretations, and we draw a theoretical comparison. Secondly, we introduce a set of uncertainty evaluation criteria, based both on existing benchmarks used in other fields and on chemistry-specific features. Finally, we implement the presented uncertainty estimation methods using as base model a recently published state-of-the-art GCNN for molecular property prediction (\texttt{chemprop}\cite{Yang2019}) and we experimentally compare them through the introduced evaluation criteria on the QM9 dataset for the regression task. In doing so, we highlight the behaviors characterizing all the methods in the context of GCNN-based molecular property prediction and their differences due to different approximation schemes. Furthermore, we discuss and quantify the positive impact of modelling uncertainty in the network on the prediction error.

\section{Methods}

\begin{figure}
    \centering 
    \includegraphics[width=1\columnwidth]{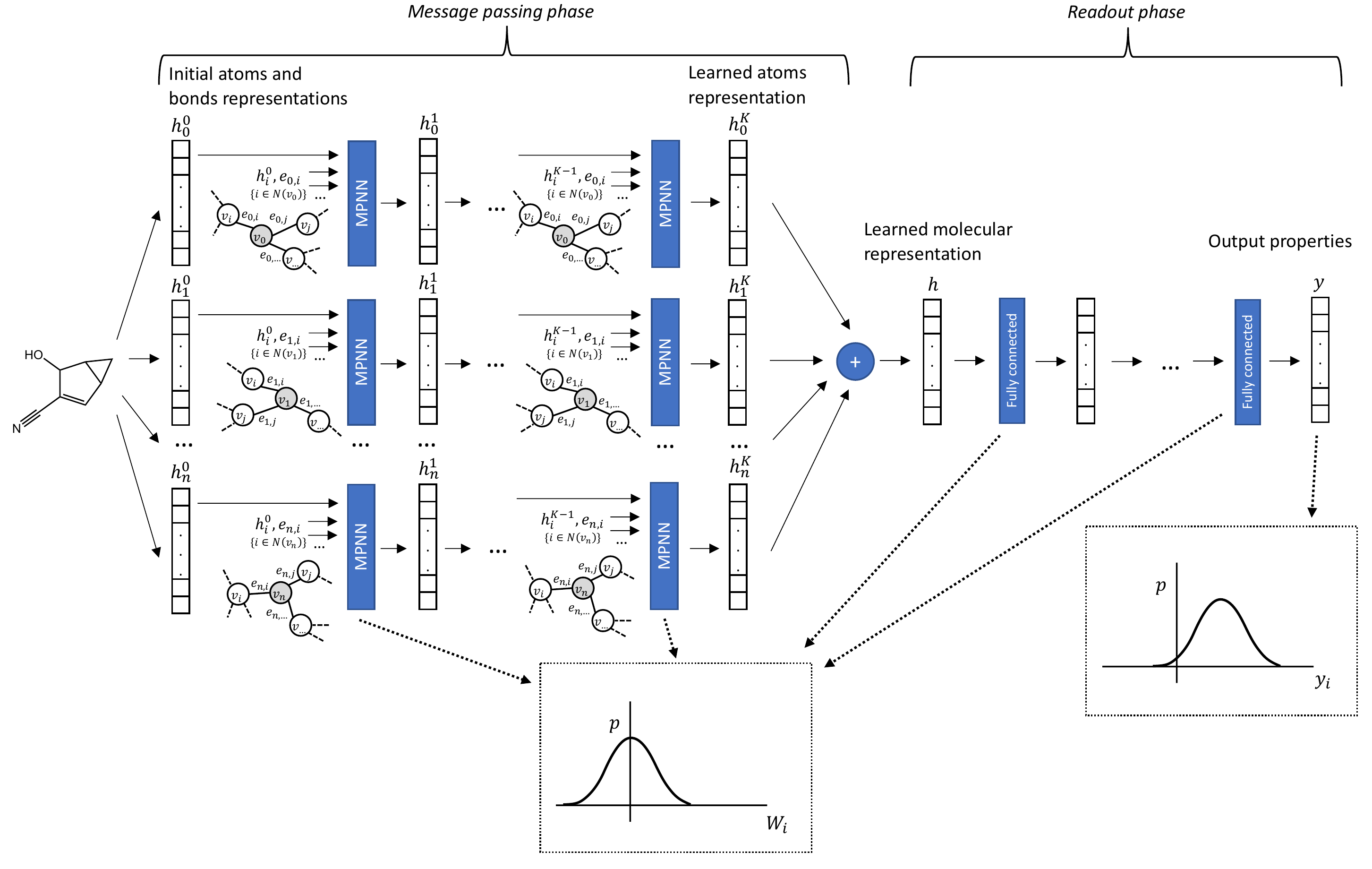}
    \caption{Illustrative overview of a GCNN for molecular property prediction extended as a BNN.
    \\ \emph{Message passing phase.} Starting from a molecular graph, the model extracts an initial representation $h_{i}^0$ for each atom $v_i$  and a bond representation $e_{i,j}$ for each bond between $v_i$ and $v_j$. At each step, the atom representation is updated based on the representation of the atom's neighbors and the related bonds. $h_{i}^j$ refers to the representation for the $i$-th atom at the $j$-th update step, $N\left({v_i}\right)$ are the atom's neighbors. At the end ($K$ update steps) the molecule representation $h$ results from the sum of the learned atoms features.
    \\ \emph{Readout phase.} The molecular representation is updated through a series of fully connected layers obtaining the output vector $y$.
    \\
    The peculiarity of a BNN is to model network weights and outputs as probability distributions (e.g., Gaussians), instead of point estimates. This allows taking into account uncertainty --- epistemic and aleatoric, respectively --- in the model.}

    \label{fig:overview}
\end{figure}

This section is organized as follows.
We first summarize GCNNs, which constitute the state-of-the-art for DNN-based molecular property prediction. We then review Bayesian Uncertainty Estimation in DNNs, detailing the methods that will be tested. Finally, we discuss uncertainty evaluation and related metrics. An overview of a GCNN extended as a DNN is shown in Figure \ref{fig:overview}.

\subsection{Graph Convolutional Neural Networks} \label{ssec:gcnn}

In general, a GCNN used for property prediction takes as input a molecular graph $G$, where the nodes are atoms and the edges are bonds, with each atom $v_i$ initialized with the feature vector $\mathbf{h}_i^0$ and each bond $v_i-v_j$ with the feature vector $\mathbf{e}_{i,j}$ and then operates in two phases (see Figure \ref{fig:overview}). During the first phase --- \emph{message passing} ---  each atom's feature  vector is updated based on the neighbors' features and related bond representations. This phase is executed $K$ times, iteratively, so that in the steps following the first one each atom's feature $\mathbf{h}_v^t$ is updated based on already updated neighbors features. This allows the interaction of distant atoms in the resulting representations. At the end, the molecule representation is given by the sum of its  atoms representations. The second phase --- \emph{readout} --- is based on a feedforward neural network that uses the final representation of the molecule to predict some properties of interest.
Intuitively, the message passing phase allows the model to learn its own feature representations directly from data, while the readout phase allows learning the relationship between such representations and output properties. The model is trained as a whole to maximize the likelihood.

Starting from this general description, several specific networks improvements have been recently proposed\cite{coley2017convolutional,wu2018moleculenet,mayr2018large,Ryu2018,Yang2019}. Given the goal of this paper of evaluating large-scale uncertainty estimation, we start from a well-tested network, \texttt{chemprop}\cite{Yang2019}, that recently reported state-of-the-art performance on multiple datasets. One of the peculiar features of this network is the usage of messages associated with directed edges (bonds) instead of vertices (atoms), improving the effectiveness of the messages exchanged. Interested readers can refer to the original work by \citet{Yang2019} for the details.

The techniques explored in this paper do not depend on a specific network, and the resulting comparative performance should hold for any GCNN model. We extended the \texttt{chemprop} model for this work to include the uncertainty estimation and evaluation methods presented next. The software developed has been made available\footnote{\url{https://github.com/gscalia/chemprop}}.

\subsection{Bayesian Uncertainty Estimation} \label{ssec:uncertainty_estimation}

Uncertainty can be the result of inherent data noise or could be related to what the model does not yet know. These two kind of uncertainties --- \emph{aleatoric} and \emph{epistemic} --- are reviewed in the next two sections, together with scalable techniques which have been proposed for their approximate computation. At the end, we discuss how these two kinds of uncertainty can be combined to obtain the \emph{total uncertainty} of a prediction. 

\subsubsection{Aleatoric Uncertainty} \label{sssection:ale_unc}

When not explicitly modeled, the inherent observation noise is assumed constant for every observed molecule. This defines a \emph{homoscedastic} aleatoric uncertainty, i.e. an uncertainty which does not vary over the data distribution and is essentially only task-dependent\cite{kendall2018multi}. However, this assumption does not hold in many realistic settings, where input-dependent noise needs to be modeled. For chemistry applications, it is usually difficult to derive a large number of high-quality data; therefore, one often needs to use multiple data sources to compose a large enough dataset to train a model. Data derived from different sources are often measured or calculated with different methods, and thus are associated with different levels of intrinsic noise. Data-dependent aleatoric uncertainty is referred to as \emph{heteroscedastic}\cite{le2005heteroscedastic} and its importance for DNNs has been recently highlighted\cite{Kendall2017}, also for molecular property prediction\cite{Ryu2018}.

\begin{figure}
    \centering 
    \includegraphics[width=0.9\columnwidth]{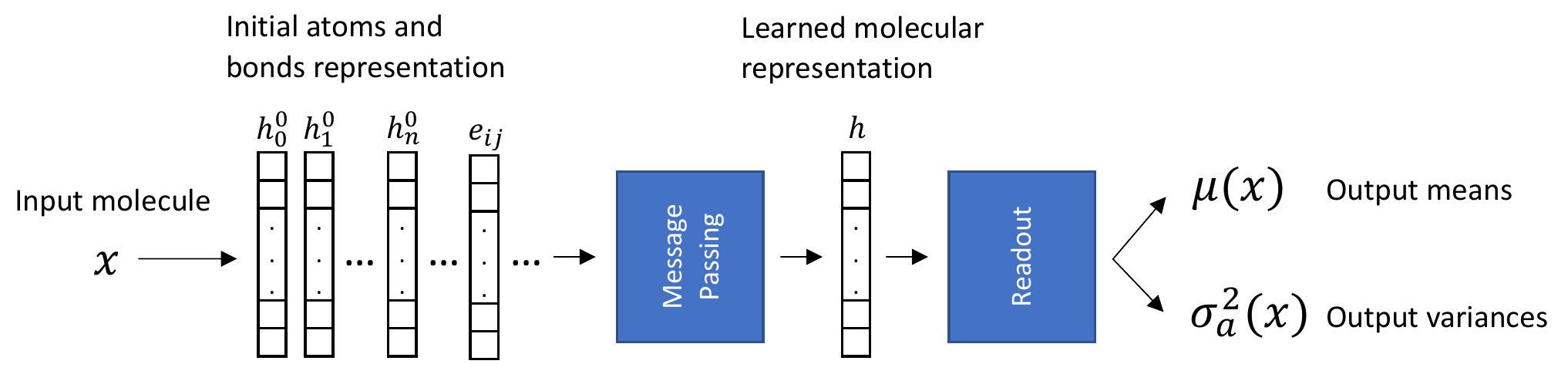}
    \caption{Aleatoric uncertainty estimation assuming an underlying Gaussian error. The last layer of the DNN is split to predict both the mean $\mu$ and the variance $\sigma$ for each output property and the network is trained minimizing the loss (Equation \eqref{eq:ale_unc}). The predicted means correspond to the output properties, the predicted variances correspond to the aleatoric uncertainties (one for each property).}
    
    \label{fig:aleatoric}
\end{figure}

Since aleatoric uncertainty is a property of data, it can be learned directly from data adapting the model and the loss function.
Assuming an underlying Gaussian error, the model (parameters $\theta$) can estimate both the mean $\mu$ and the variance $\sigma^2$ of the output distribution $\mathbf{y}$  given an input $\mathbf{x}$:

\begin{equation}
    p\left(\mathbf{y}\mid \mathbf{x},\theta\right)=\mathcal{N}\left(\mu_\theta\left(\mathbf{x}\right),\sigma^{2}_\theta\left(\mathbf{x}\right)\right)
\end{equation}

This does not require ``noise labels'' but only  changing the loss function. Indeed, by performing maximum a posteriori estimation (MAP) inference we obtain\cite{nix1994estimating}:
\begin{equation} \label{eq:ale_unc}
    \mathcal{L}\left(\theta\right) \propto \frac{1}{N}\sum_{i=1}^N\frac{1}{2\sigma^2_\theta\left(\mathbf{x}_i\right)}
    \lVert \mathbf{y_i} - \mathbf{\mu}_\theta\left(\mathbf{x}_i\right) \rVert_2^2 + \frac{1}{2}\log\sigma^2_\theta\left(\mathbf{x}_i\right)
\end{equation}
with an additional \emph{weight decay} term.
Notice that, assuming a homoscedastic uncertainty, minimizing Eq. \eqref{eq:ale_unc} coincides with the usual MSE.
In practice, the last layer of the DNN is split to predict both $\mu_\theta$ and $\sigma_\theta^2$, and the network is trained using Eq. \eqref{eq:ale_unc}, with $\sigma_\theta^2$ implicitly learned. The output $\sigma_\theta^2$ corresponds to the heteroscedastic aleatoric uncertainty: $\sigma_a^2 = \sigma_\theta^2$. This is shown in Figure \ref{fig:aleatoric}.

Interestingly, $\sigma^{2}_\theta$ in Eq. \eqref{eq:ale_unc} can be interpreted as a \emph{learned loss attenuation}\cite{Kendall2017}. Intuitively, the network can learn to increase $\sigma^{2}_\theta$ to reduce the impact of uncertain predictions on the overall loss.  The second term prevents outputting an infinite uncertainty for every point.

This approach is very practical, requiring minimal modifications to the original network, and can be used independently of the technique chosen to model weight uncertainty (epistemic uncertainty). Indeed, it has been used in conjunction  with both MC-Dropout\cite{Kendall2017} and ensembling\cite{lakshminarayanan2017simple}.

The output distribution does not need to be necessarily Gaussian (see Figure \ref{fig:overview} for a general case). In some cases, a Gaussian distribution might not be enough to model the output properties, and  more complex models could be used, such as Mixture Density Networks (MDN)\cite{bishop1994mixture}, which have been recently employed to model aleatoric uncertainty in DNN\cite{choi2018uncertainty}, or Compound Density Networks\cite{kristiadi2019predictive}, which represent a continuous extension of MDN. These  solutions allow more flexible output distributions at the cost of more complex loss functions that may  translate into less optimized and stable training. These extensions are beyond the scope of this paper.

Being predicted as a data variance, aleatoric uncertainty cannot account for uncertainty in the model's parameters $\theta$ or for other data-independent factors. Moreover, the MAP estimate does not take into account multiple plausible values for $\theta$ but only the most probable one. This can be overcome by performing Bayesian inference, as discussed next.

\subsubsection{Epistemic uncertainty} \label{sssection:epi_unc}

In a BNN the weights $\theta$ are modeled as \emph{distributions} learned from training data $\mathcal{D}$, instead of point estimates, and therefore it is possible to predict the output distribution $\mathbf{y}$ of some new input $\mathbf{x}$ through the \emph{predictive posterior distribution}, Eq. \eqref{eq:predictive_posterior_distribution}.
\begin{equation}\label{eq:predictive_posterior_distribution}
    p\left( \mathbf{y}\mid \mathbf{ x},  \mathcal{D} \right) = \int p\left( \mathbf{y}\mid\mathbf{ x}, \theta \right)p\left(\theta \mid   \mathcal{D} \right)d\theta
\end{equation}
Equation \eqref{eq:predictive_posterior_distribution} allows taking into account the epistemic uncertainty because a prediction is the ``weighted sum'' of each  outcome for each possible $\theta$ configuration of the model, with more probable $\theta$ configurations having a higher weight. The probability of a $\theta$ configuration depends on training data $\mathcal{D}$.
    
Monte Carlo integration over $M$ samples $\theta^{(i)}$ of the posterior distribution $p\left(\theta \mid \mathcal{D} \right)$ can approximate the intractable integral, however obtaining samples directly from the posterior distribution is virtually impossible for neural networks. Therefore, an approximate distribution $q\left(\theta\right) \approx p\left(\theta \mid \mathcal{D}\right)$ is introduced.

Several methods to sample from $q\left(\theta\right)$ have been introduced. The pioneering work by Neal\cite{neal1995bayesian}, employing the MCMC variant Hamiltonian Monte Carlo (HMC), is currently considered the \emph{gold standard}, but its applicability is limited to small networks and datasets. Stochastic and optimized variations have since been explored to enhance scalability at the expense of approximation performance \cite{NIPS2015_5891,zhang2019cyclical}.

\emph{Variational Inference} (VI) is an alternative paradigm to derive $q\left(\theta\right)$. In this case, a class of approximating distributions $q_\phi\left(\theta\right)$ parameterized by $\phi$ is explicitly chosen, so that posterior approximation becomes an optimization problem of finding $\phi$ miniziming the Kullback-Leibler (KL) divergence with respect to $p\left(\theta \mid \mathcal{D}\right)$. The set of approximating distributions is pre-defined and performance will depend on the search space and the employed optimization procedure.

VI methods constitute a standard technique in Bayesian modelling. However, scalability requirements and NN-specific features have led to the design of new methods for this class of models in the last few years\cite{graves2011practical,hernandez2015probabilistic,gal2016dropout,duvenaud2016early,liu2016stein}. Nonetheless, some of these approaches --- such as Stein Variational Gradient Descent\cite{liu2016stein} --- do not actually scale up to training-intensive applications such as active learning based molecular property prediction\cite{Zhang2019}.

MC-Dropout and ensembling-based methods are currently the most popular approaches for large-scale uncertainty estimation in NNs\cite{gustafsson2019evaluating} and, within chemistry, both have been very recently introduced\cite{Ryu2018,Zhang2019,Li2019,smith2018less,peterson2017addressing}. In addition to their scalability, these methods owe their popularity to the relative ease of implementation, since both leverage well-known techniques for regularization and accuracy improvement. 
For this reason, in the following we will focus on MC-Dropout and ensembling, describing both the original methods, main variations (in particular, bootstrapping), recent improvements and interpretations.

\paragraph{Monte Carlo Dropout}

MC-Dropout\cite{gal2016dropout,Kendall2017} is a simple and scalable VI approach. The algorithm consists in training a network with  dropout before every layer and then, at testing time, keeping dropout to sample $M$ outputs $\mathbf{y}^{(i)}$ with different random masks. Each different random dropout mask  corresponds to a sample from the approximate posterior $q_\phi\left(\theta\right)$. The model prediction $\mathbf{\Tilde{y}}$ is the mean of the different outputs, while the epistemic uncertainty $\sigma_e^2$ can be captured by the variance of the different outputs. If the aleatoric uncertainty is also computed (as in Figure \ref{fig:aleatoric}), the output aleatoric uncertainty is the mean of the different aleatoric uncertainty estimates (and, in this case, the $\mathbf{y}^{(i)}$ are substituted by the $\mathbf{\mu}^{(i)}$):
\begin{equation} \label{eq:mean_variance}
    \mathbf{\Tilde{y}} = \frac{1}{M}\sum \mathbf{y}^{(i)} \quad \sigma_e^2 = var\left(\mathbf{y}^{(i)}\right) \quad \sigma_a^2 = \frac{1}{M}\sum \sigma_a^{(i)}
\end{equation}

Formally, the MC-Dropout algorithm approximates the posterior with a product of Bernoulli distributions. Indeed, given a dropout probability $p$, each unit of the network with parameters $\theta_i$ has probability $p$ of being dropped and set to zero. Equivalently, the approximation distribution can be seen as a mixture of two Gaussians with small variances and the mean of one of the Gaussians is fixed at zero\cite{gal2016dropout,Kendall2017}.

A drawback of the MC-Dropout approach is the introduction of the dropout rate $p$ as hyper-parameter. Such a choice has an important impact both on the model's accuracy and the uncertainty estimation. Indeed, $p$ contributes to determine the magnitude of the epistemic uncertainty.
Moreover, this hinders model hyper-parametrization, especially if $p$ is chosen to be layer-dependent.

Among the methods proposed in the literature to automatically tune the dropout probability, Concrete Dropout\cite{gal2017concrete} represents a practical gradient-based solution which follows dropout’s variational interpretation. This approach has demonstrated comparable performance with respect to grid-searched $p$\cite{gal2017concrete} and an improvement in model calibration with respect to standard MC-Dropout\cite{mukhoti2018evaluating}.
Therefore, we will compare this non-parametric version of MC-Dropout to the intrinsically non-parametric ensembling approach.

\paragraph{Ensembling}
Ensembling has been introduced as a practical non-Bayesian alternative to estimate uncertainty in \citet{lakshminarayanan2017simple}. The algorithm consists in training the same network multiple times with a random initialization, minimizing the MLE objective $-\log p\left( \mathbf{y}\mid \mathbf{x}, \theta\right)$ each time. The output of the ensemble is given by the mean of the predictions, while the variance corresponds to the ensemble uncertainty, as in Equation \eqref{eq:mean_variance} for MC-Dropout.

It is possible to draw a parallel between ensembling and MC-Dropout, since the latter can also be interpreted as a form of ensembling \cite{lakshminarayanan2017simple, srivastava2014dropout} with weight sharing between the models.
Even if ensembling has been originally proposed as a non-Bayesian solution \cite{lakshminarayanan2017simple}, recent literature has proved how, with minor modifications to the original ensembling methodology, it is possible to interpret it as a Bayesian inference technique \cite{duvenaud2016early, pearce2018uncertainty}. Nonetheless, even without the modifications, ensembling can be interpreted as Bayesian approximation with an implicit distribution $q\left(\theta\right)$  \cite{gustafsson2019evaluating}.

Ensemble methods have long been recognized as very effective to improve predictive performance of machine learning \cite{dietterich2000ensemble} and deep learning models \cite{Goodfellow2016}, and their effectiveness for this purpose has been assessed even recently in chemistry for QSPR \cite{Yang2019}.
The reason why ensembling allows reducing the overall error with respect to each of components resides in the diversity of their errors. Indeed, perfectly correlated errors do not bring any advantage to the ensemble error, while perfectly uncorrelated errors reduce the expected ensemble error proportionally to the number of employed instances \cite{Goodfellow2016}. Different solutions can be easily reached by deep models given their nonconvexity and the sub-optimal optimization strategies employed.

The intuition behind the interpretation of the ensemble variance as model uncertainty is simple. Different instances of the ensemble of models will tend to output similar values when the inputs are similar to the observed training data, because each instance's weights, even if different, are optimized for those data. In contrast, as inputs become less similar to the training data, the outputs of each instance tend to be more affected by the specificities of the sub-optimal solution reached, thus the higher variance.
Given this, it seems clear that diversity in the ensembled models should be promoted both for error reduction and uncertainty improvement.

Traditional regularization techniques, such as weight decay and early stopping, affect the solutions reached by NNs.  Recently, the usage of these techniques has been proposed not only as a practical strategy to increase ensemble diversity, but also as a formal evidence for a Bayesian interpretation of ensembling\cite{pearce2018uncertainty,duvenaud2016early}. This is discussed in the next paragraph.

\paragraph{Anchored Ensembles and early stopping}

Anchored ensembling \cite{pearce2018uncertainty} modifies traditional ensembling leveraging the \emph{randomised MAP sampling} technique. This technique exploits the fact that injecting some noise in the loss function of a MAP estimate allows sampling from the true posterior. Therefore, an ensemble of such models is a simple and scalable approach for approximate Bayesian inference.

It is known that the commonly used $L_2$ regularization for NN (weight decay) corresponds to the MAP estimate with Gaussian priors\cite{Goodfellow2016}, which can be interpreted as pulling the weights for which the network does not express a strong preference close to zero.
The anchored ensembling algorithm proposes to add noise to this loss function by changing the priors' means. For regression, this leads to the following loss for the $i$-th model  in the ensemble:
\begin{equation}
    \mathcal{L} = \frac{1}{N}\lVert \mathbf{y} - \mathbf{\Tilde{y}}^{(i)} \rVert_2^2 + \frac{1}{N} \lambda\lVert \mathbf{\theta}^{(i)} - \mathbf{\theta}_{0}^{(i)} \rVert_2^2
\end{equation}
where $\mathbf{y}$ are the target outputs and $\mathbf{\theta}_{0}^{(i)}$, which equals to zero for standard $L_2$ regularization, is the prior's mean of the $i$-th model.

Following this approach, each model in the ensemble has its parameters \emph{anchored} to a different $\mathbf{\theta}_{0,i}$, and this promotes the diversity of the solutions reached by the different models. 

An important limitation of this approach is the need for additional hyper-parameters that must be tuned. They include at least the regularization coefficient $\lambda$ --- that expresses the ratio between data variance and weights' prior variance --- and the noise distribution $\mathbf{\theta}_{0,i} \sim \mathcal{N}\left(0,\Sigma_0\right)$. As originally described\cite{pearce2018uncertainty}, the algorithm also employs a regularization matrix $\mathbf{\Gamma}$ instead of the scalar $\lambda$, to allow specifying per-layer regularization. 

The work presented in \citet{duvenaud2016early} gives an interesting interpretation to a commonly exploited regularization method --- \emph{early stopping} --- as approximate nonparametric Bayesian VI. In particular, they show how training a model to minimize the negative log-likelihood with stochastic gradient descent (SGD)\footnote{The approach is compatible also with minibatches.} can be interpreted as obtaining the approximate posterior $q_t\left(\theta\right)$ parametrized by the number $t$ of SGD steps, and demonstrate how early stopping leads to an optimal $\Tilde{t}$.  Within this context, the initial distribution of the model $p\left(\theta_0\right)$ is interpreted as the \emph{prior}.

In practice, $q_{\Tilde{t}}\left(\theta\right)$ allows sampling from the variational posterior, and therefore ensembling different random restarts allows obtaining independent samples from the posterior, that can then be used as in traditional ensembling (Eq. \eqref{eq:mean_variance}). Even if the approach, as originally described, does not take into consideration SGD with momentum, recent work also shows how SGD with momentum can be interpreted as Bayesian inference \cite{mandt2017stochastic}.

Not only is this approach practical, but ensembling with early stopping is  usually already exploited for property prediction in state-of-the-art systems \cite{Yang2019}. In this work we use it as a Bayesian alternative for uncertainty estimation.

We can draw a parallelism between the two approaches described above. It has been shown that early stopping for NNs is conceptually similar to $L_2$ regularization, while an exact equivalence holds in the simpler case of a linear model with a quadratic loss function\cite{Goodfellow2016}. Intuitively, both approaches restrict the optimization procedure to the vicinity of a pre-defined value --- $\theta_0$ for $L_2$ regularization, the initial configuration for early stopping. In our case, we notice that these two values have the same role of \emph{prior} in the two approaches\cite{pearce2018uncertainty,duvenaud2016early}, highlighting an interesting similarity. Even though they are based on different theoretical foundations, in practice both the approaches increase the diversity in the ensembled instaces by injecting some randomness into their regularization. 
An intrinsic advantage of early stopping over weight decay is that early stopping automatically determines the correct amount of regularization, instead of requiring external hyper-parameter optimization\cite{Goodfellow2016}. Therefore, given the objective of this paper of evaluating scalable and practical uncertainty quantification techniques, in the following we will focus on early stopping for our extensive tests. Anchored ensembling and the impact of different priors for uncertainty estimation will be the subject of future work.

\paragraph{Bootstrapping}

Also referred to as bagging, bootstrapping is a popular technique where ensemble members, instead of being trained on the whole dataset, are trained on different \emph{bootstrap samples} of the original training set. Each bootstrap sample $\mathcal{D}_i$ is obtained by sampling $K$ samples with replacement from the dataset $\mathcal{D}$ and therefore will include a fraction of the elements in $\mathcal{D}$  and duplicates. If the original dataset is a good  approximator of the underlying distribution, each $\mathcal{D}_i$ will also be.

Bootstrapping allows increasing the diversity in the trained instances, which, as previously discussed, is a key factor for ensembling performance. However, instead of relying on diversity in the models, bootstrapping relies on  diversity in the datasets.

This approach has been successfully employed to increase the diversity in shallow ensembles, but its use within NNs might be less beneficial, since, given the dependence on a large amount of training data, each individual instance will be less powerful, thus affecting the whole ensemble performance\cite{lakshminarayanan2017simple}. Moreover, recent progresses in NN understanding suggest these models are characterized by an extremely large amount of equivalent local minima\cite{Goodfellow2016}, and the inherent stochasticity of SGD should already provide some degree of diversity even when trained on the same dataset.

Nonetheless, since bootstrapping has been recently described in the literature as an effective approach for NNs\cite{peterson2017addressing,Li2019}, we aim to compare it to full-dataset ensemble in different operating conditions to assess the differences with respect to the various evaluation metrics introduced.

\bigbreak

\begin{figure}
     \centering
     \begin{subfigure}[b]{0.45\textwidth}
         \centering
         \includegraphics[width=\textwidth]{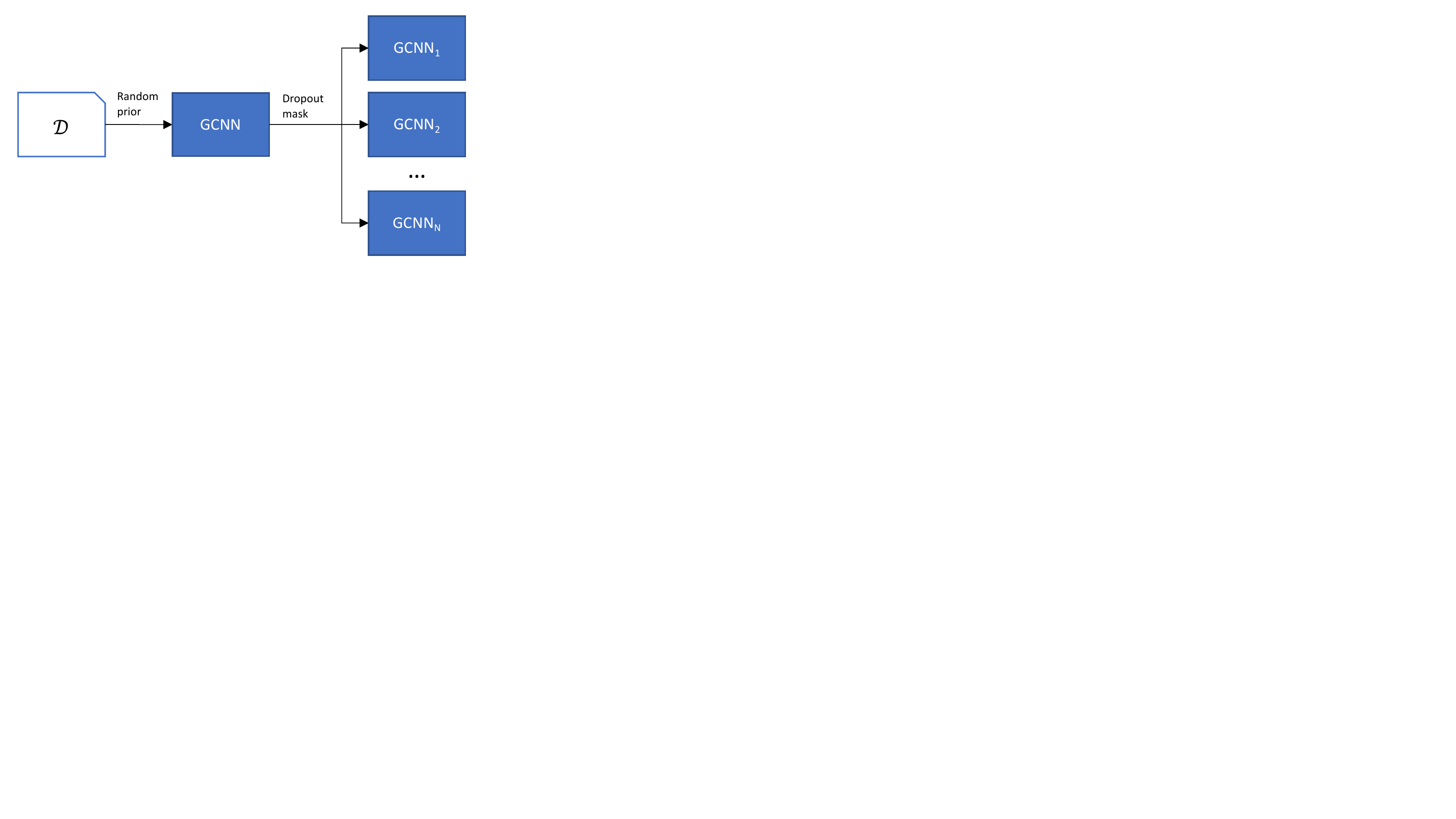}
         \caption{\textbf{MC-Dropout}. Only a network is trained to minimize the loss on the training dataset. Then, at testing time, multiple models are ``generated'' applying a stochastic dropout mask to the initial network. All the models \textsf{GCNN$_1$},\ldots \textsf{GCNN$_N$} share (part of) the same weights. Diversity in the models is the result of dropout masks.}
         \label{fig:overview-mc_dropout}
     \end{subfigure}
     \hfill
     \begin{subfigure}[b]{0.30\textwidth}
         \centering
         \includegraphics[width=\textwidth]{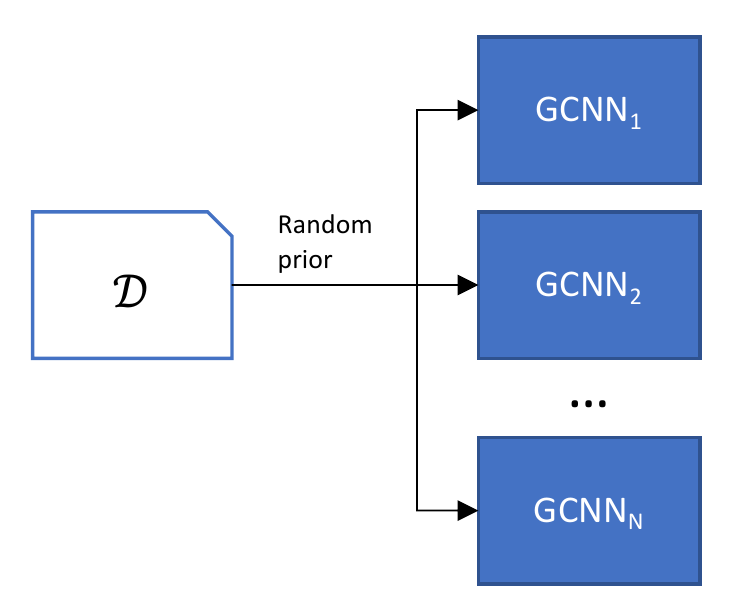}
         \caption{\textbf{Ensembling}. Different models are trained to minimize the loss on the same training dataset. Diversity in the models results from different initial configurations (random priors).}
         \label{fig:overview-ensembling}
     \end{subfigure}
     \hfill
     \begin{subfigure}[b]{0.45\textwidth}
         \centering
         \includegraphics[width=\textwidth]{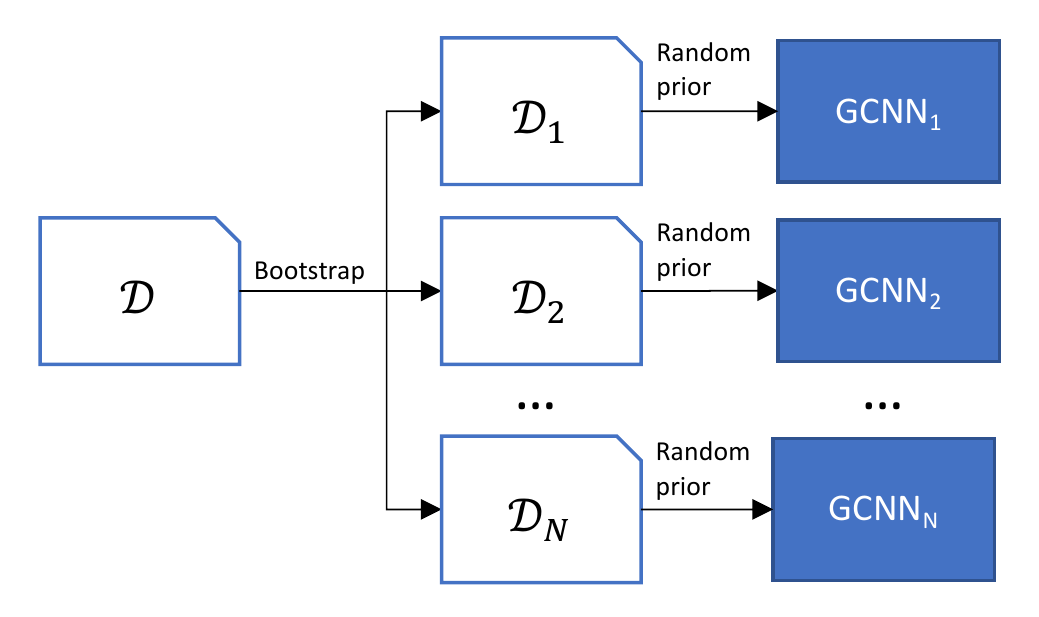}
         \caption{\textbf{Bootstrapping}. Each model is trained to minimize the loss on a \emph{bootstrap sample} of the training dataset. This, together with different initial configurations (random priors) ensures diversity in the models.}
         \label{fig:overview-bootstrapping}
     \end{subfigure}
        \caption{Overview and comparison of MC-Dropout, ensembling and bootstrapping. $\mathcal{D}$ is the training dataset.}
        \label{fig:overview-models}
\end{figure}

\begin{figure}
    \centering 
    \includegraphics[width=0.55\columnwidth]{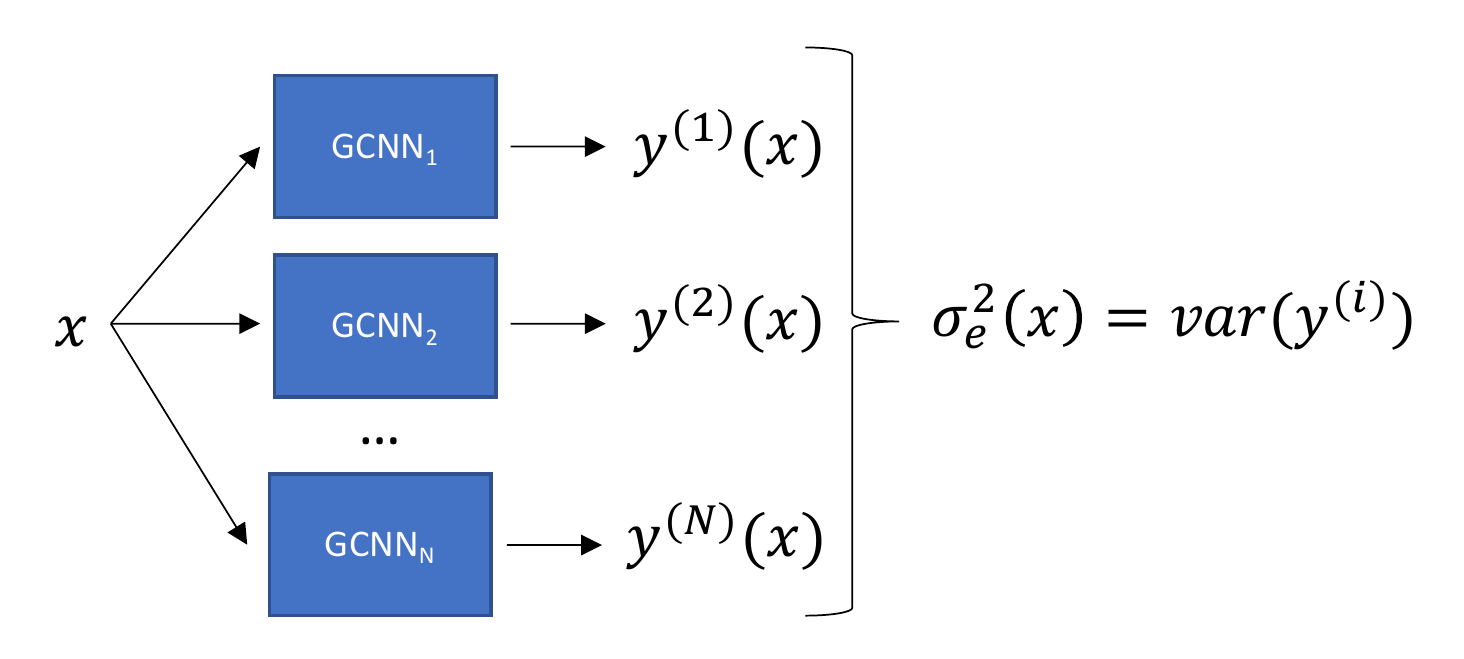}
    \caption{Epistemic uncertainty computation. Independently from the method used to obtain multiple models (see Figure \ref{fig:overview-models}), the epistemic uncertainty is estimated as the variance of the different outputs (ref. Eq. \ref{eq:mean_variance}). If the aleatoric uncertainty is also computed (as in Figure \ref{fig:aleatoric}), the output aleatoric uncertainty would be the mean of the different aleatoric uncertainty estimates (and, in this case,  $\mathbf{y}^{(i)}$ would be substituted by $\mathbf{\mu}^{(i)}$).}
    
    \label{fig:epistemic}
\end{figure}

A comparative overview of MC-Dropout, ensembling and bootstrapping is presented in Figure \ref{fig:overview-models}. As shown, each method relies on a set of predictions (explicit or implicit models), which diversity is driven by different factors. The different predictions are used to estimate epistemic uncertainty as shown in Figure \ref{fig:epistemic}.

\subsubsection{Total uncertainty} \label{sssec:total_uncertainty}
Aleatoric and epistemic uncertainty can be added to approximate the \emph{total uncertainty} of a prediction\cite{gal2016uncertainty,Kendall2017}.
The total uncertainty captures all the variability of the output $\mathbf{y}$, which includes both the variability coming from our ignorance about the model (epistemic uncertainty) and variability coming from inherent randomness of the output (aleatoric uncertainty). We will evaluate both the separate contributions and the total uncertainty.

\subsection{Uncertainty Evaluation} \label{ssec:evaluating_uncertainty}

In the following, several methods to evaluate the accuracy of uncertainty estimates are discussed. We start from existing techniques described in the literature, merging the contributions of different fields, and we extend them to account for  specific features of chemical space.
We aim at identifying a set of quantitative and complementary evaluation criteria. 
First, we introduce \emph{ranking based methods}, i.e. evaluation criteria based on the uncertainty's capability of ordering predictions based on their confidence. 
Secondly, we discuss \emph{calibration}, i.e. ``the property of predicting probability estimates representative of the true correctness likelihood''\cite{guo2017calibration}. 
Then, \emph{dispersion} is introduced to complement  calibration evaluation. 
Finally, we discuss uncertainty \emph{domain shift}, i.e. the property of predicting reliable uncertainty estimates for molecules different with respect to those seen during training.

\subsubsection{Ranking based methods} \label{sssec:ranking_methods}

A first class of evaluation indexes is based on the ranking defined by uncertainty estimates. This  allows defining a \emph{confidence curve}, which, in turn, allows defining several quantitative indices.

\paragraph{Confidence curve}
One way to evaluate the uncertainty is by considering how the error varies as we remove  molecules with the highest uncertainty in the test dataset. Indeed, a meaningful uncertainty should lead to a lower error on a subset of high-confident predictions.
This concept is captured by the \emph{confidence curve}, that highlights how the error varies (with respect to a given metric, e.g. MAE or RMSE) as a function of  confidence percentile (or, in general, confidence $q$-quantile), i.e. the error on the subset of \emph{n}\% molecules (\emph{n}-th $q$-quantile) with the lower uncertainty.

Ideally, we would expect a decreasing confidence curve for a meaningful uncertainty. The error corresponding to the left-most point is simply the error on the complete test dataset; the following points correspond to the error on the subset of testing molecules belonging to the \emph{n}-th $q$-quantile.
Other than being decreasing, another important feature of the confidence curve is its shape: a better uncertainty corresponds to a higher slope,  because it allows decreasing the error faster for the same amount of removed molecules. For comparison, randomly sampling the molecules to be removed should lead to a more or less constant function.

What this kind of evaluation really assesses is the \emph{ordering} of the predictions by their confidence.
From this perspective, the best possible ordering is the one imposed by the true error, which has been named \emph{oracle} ordering\cite{ilg2018uncertainty} in the literature. We can interpret the oracle ordering as an \emph{uncertainty lower bound}, and  the \emph{oracle confidence curve} is the best confidence curve obtainable for a given model and test data.

\paragraph{Confidence-Oracle error and AUCO}

Since the oracle ordering corresponds to the lower bound, we can define the \emph{Confidence-Oracle error} as the difference between the confidence curve for a given uncertainty estimation, $h^{(i)} = \left(h^{(i)}_1, h^{(i)}_2, \ldots h^{(i)}_{q-1}\right)$ and oracle confidence curve, $h^{(o)}= \left(h^{(o)}_1, h^{(o)}_2, \ldots h^{(o)}_{q-1}\right)$.
In general, we want this error to be as small as possible, therefore we introduce the \emph{Area Under the Confidence-Oracle error}, AUCO, to quantify it in a single number
\footnote{
The Confidence-Oracle error has been called \emph{Sparsification Error} in the context of optical flow estimation in computer vision\cite{ilg2018uncertainty}.
The AUCO has been called Area Under the Sparsification Error curve in the same context\cite{ilg2018uncertainty}.}:
\begin{equation}
    \text{AUCO}\left(h^{\left(i\right)}\right) = \sum_{j=1}^{q-1}\left(h^{(i)}_j - h^{(o)}_j\right)
\end{equation}
This value allows an easy comparison between two uncertainty estimations $h^{(i)}$ and $h^{(j)}$ with respect to the oracle, where the smaller is better.

For this kind of comparison, it is important to highlight that every confidence curve depends not only on the uncertainty estimation, but also on the predictive model. Indeed, while the first defines the $q$-quantiles, the second provides the data for which each quantile error is calculated. It follows that it is not possible to directly compare two confidence curves obtained through different models to establish which uncertainty estimation is better. This is particularly relevant because often the uncertainty estimation and the predictive model are strongly tied: for example, ensembling is an uncertainty technique that also affects the predictive model.

With this regard, an added benefit of the confidence-oracle error is that, since it marginalizes out the oracle, it enables a fair comparison of uncertainty estimates based on different methods \cite{ilg2018uncertainty}. Therefore, the confidence-oracle error and the AUCO will be used in the following for this purpose.

Notice that, using $q$-quantiles, each uncertainty-imposed ranking that does not change the specific quantile each prediction belongs to, even if it does change the relative position of the $q$ predictions inside each quantile, is equivalent from the point of view of the confidence curve, the confidence-oracle error and the AUCO. Hence it follows that these are all affected by the choice of $q$. 
In the following, we will use percentiles as commonly reported in the literature. 

\paragraph{Error Drop}
As  an additional quantitative measure of confidence curve quality that does not depend on the oracle, we introduce the \emph{Error Drop}. 
This is defined  as the error ratio between the first and last quantiles, which should correspond to the curve's maximum and minimum, respectively, if the confidence curve behaves correctly:
\begin{equation}
    \text{Error Drop}\left(h^{\left(i\right)}\right) = \frac{h^{(i)}_1}{h^{(i)}_{q-1}}
\end{equation}

This index measures the relative performance improvement of the model obtainable by considering only the most confident predictions instead of the entire dataset. Being a ratio, we can use it to directly compare different methods.  

\paragraph{Decreasing coefficient}
A limitation of the AUCO and Error Drop indices is that they do not take into account the monotonicity of the confidence curve. We observe that in existing evaluations this property is usually qualitatively considered but not quantitatively measured, and therefore we introduce a \emph{Decrease Ratio} to capture it. Given a confidence curve $h^{(i)} = \left(h^{(i)}_1, h^{(i)}_2, \ldots h^{(i)}_{q-1}\right)$:
\begin{equation}
    \text{Decr. Ratio}\left(h^{\left(i\right)}\right) = \frac{\left\lVert \left\{ h^{(i)}_j \mid h^{(i)}_j\geq h^{(i)}_{j+1} \right\} \right\rVert}{q-1} \quad \forall j \in 1, \ldots, q-1
\end{equation}
where $\text{Decr. Ratio} = 1$ corresponds to a perfectly non-increasing curve.

Rather than being a measure of uncertainty quality by its own, this coefficient captures the noise in the confidence curve and should be used in combination with the other metrics for a more comprehensive analysis. 

\subsubsection{Uncertainty Calibration} \label{sssection:calibration}
One limitation of the evaluation methods introduced up to now is that they are all order-based, and therefore they only take into account the ranking imposed by uncertainty estimates and true errors. While this is crucial to distinguish among various degrees of model confidence, it does not take into consideration the \emph{actual values} expressed by uncertainty.

Indeed, another important aspect of uncertainty is more strictly related to the actual values it expresses, and referred to as \emph{calibration}.
In general, calibration of a model refers to the property of outputting probability distributions which are consistent with observed empirical frequencies. 

Calibration evaluation of neural networks gained interest in the last two years, since it has been shown that modern neural networks, while being more accurate on one side, are less calibrated on the other\cite{guo2017calibration}, thus encouraging more research on the topic\cite{Kendall2017,lakshminarayanan2017simple}. Indeed, model calibration is orthogonal with respect to model accuracy\cite{lakshminarayanan2017simple}.  Calibrated confidence is important for model interpretability and to establish trustworthiness with the user\cite{guo2017calibration}, since it allows providing  uncertainty estimates which are informative not only relatively to other estimates, but also on their own with respect to model's predictions. 

Model calibration can be easily defined in the classification setting, since, given an input $\mathbf{x}$, an output $H\left(\mathbf{x}\right)$ and a vector confidence $\mathbf{h}$ over the set of classes $\mathcal{C}$, the model is considered perfectly calibrated when the following holds:
\begin{equation}
    p\left(H\left(\mathbf{x}\right) = c \mid \mathbf{h}_c = \Tilde{p}\right) = \Tilde{p},\quad \forall \Tilde{p} \in \left[0,1\right]
\end{equation}
where $\mathbf{h}_c$ is the confidence associated to the class $c \in \mathcal{C}$. This means that the confidence assigned to each class is consistent with the probability of a prediction of belonging to that specific class.

In practice, over a finite number of samples, calibration can be captured by a \emph{Calibration Plot}\cite{Kendall2017}, also called \emph{Reliability Diagram}\cite{guo2017calibration}.
To obtain such a plot the model predictions for all samples and classes in the test set are split into $K$ \emph{bins}\footnote{Each bin is a subset of predictions.} in the range $\left[0,1\right]$ and the frequency of correctly predicted labels for each bin is plotted\cite{niculescu2005predicting}. Perfect calibration corresponds to a \emph{diagonal line}.

Calibration can vary within the same uncertainty estimator when considering different uncertainty intervals. This could happen, for example, if a model has well-calibrated low uncertainty but ill-calibrated high uncertainty, or vice-versa. Such cases are highlighted by a Calibration Plot which diverges from the diagonal line in some specific confidence intervals but not in others.

\paragraph{Calibration in regression}
Uncertainty calibration is a well-studied topic in the context of classification,  both in its traditional  domain of weather forecasting\cite{degroot1983comparison} and, more recently, in deep learning\cite{guo2017calibration}. However, calibration for regression  appears to be less investigated, and different solutions to evaluate it have been employed and discussed only recently\cite{Kendall2017,gustafsson2019evaluating,kuleshov2018accurate,levi2019evaluating}.
Focusing on molecular property prediction, calibration for regression becomes crucial to account for scalar properties like formation enthalpies or energies.
In the following, we will consider two different definitions which extend calibration in a regression setting: \emph{confidence-intervals} based and \emph{error} based calibration.

\begin{itemize}
    \item \textbf{Confidence-based calibration} (also called interval-based calibration)\cite{kuleshov2018accurate,gustafsson2019evaluating} interprets each prediction and its uncertainty as the mean and the variance of a Gaussian distribution $p\left(\mathbf{y}\mid \mathbf{x}\right)=\mathcal{N}\left(\mu\left(\mathbf{x}\right),\sigma^{2}\left(\mathbf{x}\right)\right)$, respectively, and we are interested in evaluating the confidence intervals thus defined. To do so, we consider symmetric intervals of varying confidence around the mean and compare them to the empirical probabilities of belonging to each interval. In a well-calibrated model, the $x$\% of the predictions should fall in the $x$\% confidence interval.
    In practice, we discretize the confidence intervals and calculate the fraction of predictions falling in each interval. This allows obtaining a Calibration Plot in the  $\left[0,1\right]$ range, as in the classification case, where perfect calibration corresponds to a diagonal line.

    \item \textbf{Error-based calibration}, originally described by \citet{levi2019evaluating}, proposes to directly compare the uncertainty to the empirical error, as in Eq. \eqref{eq:error_calibration}.
    
    \begin{equation} \label{eq:error_calibration}
        \EX\left[\left(\mu\left(\mathbf{x}\right) - \mathbf{y}\right)^2 \mid \sigma^2\left(\mathbf{x}\right) = \Tilde{\sigma}\right] = \Tilde{\sigma},\quad \forall \Tilde{\sigma}
    \end{equation}
    
    This defines a perfectly calibrated model as one outputting an uncertainty matching the expected error. As in the classification case, in practice, to assess  calibration it is necessary to split the test data ordered by estimated uncertainty in $K$ bins and average uncertainties and errors for each bin.
    It is then possible to define the Calibration Curve by plotting the MSE of the $i$-th bin as a function of its average uncertainty
    \footnote{In the original definition proposed in \citet{levi2019evaluating}, the RMSE and the predicted standard deviations are used instead of MSE and  variances. We use the latter for consistency with the other measures introduced.}.
    Notice that, unlike classification and confidence-interval calibration cases, here the Calibration Plot is not bound in the $\left[0,1\right]$ interval but ranges between 0 and the maximum  uncertainty.
    As in the other cases, perfect calibration corresponds to a diagonal line.
\end{itemize}

Each of these two approaches has its pros and cons. Confidence-based calibration has the advantage of considering all the predictions to compute  each point of the plot, thus resulting in more robust empirical calculations. However, as recently highlighted\cite{levi2019evaluating}, one can re-calibrate practically any output distribution using this evaluation method --- even an entirely uncorrelated uncertainty. While this is not a limitation for the present work, since we do not address uncertainty re-calibration, it is something to be taken into consideration in general.
The main advantage of error-based calibration is that it directly ties computed uncertainty to expected error, thus reflecting what the user would expect. The main limitation is represented by the fact that, since only a fraction of uncertainty estimates contributes to each computed point, and the uncertainty estimates are not uniformly distributed, the  subsets used to compute the different points are not homogeneous. 

Independently from which method is used to form a Calibration Plot, it is then possible to define some metrics over it to quantify calibration performance, as discussed in the next paragraphs.

\paragraph{Calibration Error Curve and AUCE}
We can evaluate uncertainty calibration by computing the absolute difference of the Calibration Plot with respect to perfect calibration, thus obtaining the \emph{Calibration Error Curve}. This difference can be  quantified by considering the area under this curve, which has been referred to as the \emph{Area Under the Calibration Error Curve}, AUCE metric\cite{gustafsson2019evaluating}. 
This is a cumulative metric accounting for the total calibration error.

\paragraph{ECE, MCE and ENCE}
Rather than considering the total error, it is possible to define the \emph{Expected Calibration Error} (ECE) and the \emph{Maximum Calibration Error} (MCE) as follows (for the simpler binary classification case)\cite{naeini2015obtaining,guo2017calibration}:
\begin{align}
\begin{split}
    ECE = \sum_{i=1}^K p\left(i\right) \cdot \lVert \text{acc}\left(b_i\right) - \text{conf}\left(b_i\right) \rVert
    \\ 
    MCE = \max_{i=1}^K \left( p\left(i\right) \cdot \lVert \text{acc}\left(b_i\right) - \text{conf}\left(b_i\right) \rVert \right)
\end{split}
\end{align}
where $b_i$ is the $i$-th bin, $p\left(i\right)$ is the fraction of predictions that fall into the bin, $\text{acc}$ and $\text{conf}$ are the accuracy (i.e., the fraction of times a class is correctly predicted) and the average confidence for the bin. ECE and MCE correspond to the average and the maximum over the Calibration Error Curve, respectively, weighted by the fraction of predictions which contribute to each bin. MCE is especially important in high-risk applications, since it models the \emph{worst-case scenario}\cite{guo2017calibration}. 

This definition can be extended for regression. 
For confidence-intervals based calibration we can compare the prediction accuracy (i.e. the fraction of times a prediction falls into the confidence interval) to the confidence. In this case $p\left(i\right)=\frac{1}{K}$ since all the predictions contribute to all the bins.
For error-based calibration $\text{acc}$ and $\text{conf}$ are substituted by the RMSE and the root mean uncertainty, respectively, and this discrepancy is further normalized by the uncertainty  over the bin, since the error is expected to be naturally higher as the uncertainty increases\cite{levi2019evaluating}, thus defining the Expected Normalized Calibration Error (ENCE). 

\subsubsection{Sharpness and dispersion} \label{sssec:sharpness}

Calibration by itself could be insufficient to fully evaluate an uncertainty estimator. Indeed, if the model always outputs the \emph{same constant uncertainty} which matches the empirical accuracy over the entire distribution, we obtain a perfectly calibrated uncertainty but not a very useful one, since it does not depend on the input data at all.
This concept is captured by \emph{sharpness}, an uncertainty's property orthogonal and complementary to calibration \cite{gneiting2007probabilistic}. Originally defined in the classification settings, it intuitively refers to outputting probabilities which are as much as possible concentrated around specific classes (for example, in a binary setting, probabilities close to zero or to one). From another perspective, it rewards \emph{input-dependent} uncertainty estimates.

This notion has been recently extended for regression\cite{kuleshov2018accurate,levi2019evaluating}. Following the definition introduced in \citet{levi2019evaluating}, in the following the \emph{dispersion} of an uncertainty estimator is defined as the \emph{coefficient of variation} $c_v$ of its uncertainty estimates (interpreted as standard deviations).
A higher $c_v$ corresponds to more heterogeneous estimates for different inputs.

It should be noted that, for different reasons, dispersion cannot be used as an absolute measure to quantify the performance of a given uncertainty estimator on a given dataset. First of all, a higher $c_v$ by itself does not necessary reflect into more accurate confidence estimates. Secondly, the ``true'' dispersion depends on the dataset and could also be naturally low for homogeneous datasets. Moreover, being a normalized measure, $c_v$ does not take into consideration the absolute uncertainty values but only their dispersion around the mean.
Nonetheless, dispersion represents a useful metric to be taken into account along with calibration when comparing different methods. In particular, we are interested in verifying that an improvement in calibration of an uncertainty estimator with respect to another one does not originate from a reduction in  dispersion.

To the best of our knowledge dispersion has not been taken into account before in comparative evaluations of deep learning uncertainty estimation frameworks \cite{guo2017calibration,ilg2018uncertainty,gustafsson2019evaluating,beluch2018power} or in the context of deep molecular property prediction\cite{Ryu2018,Zhang2019}, thus further motivating its experimental evaluation in the following.

\subsubsection{Domain shift} \label{sssec:domain_shift}

An important feature that should characterize a well-behaving uncertainty estimate is its ability to correctly manage \emph{domain shifts}, i.e., its performance in an \emph{out-of-domain} context, which corresponds to a test set that is markedly different to the one seen during training. While this behavior --- which implies a low variance of the model --- is of first importance for every model's output, it becomes even more crucial for uncertainty estimates. Indeed, it is well known that every learned model will degrade at some point on unseen samples as they become more and more different with respect to those seen during training,  but a well-calibrated uncertainty should be able to correctly identify this ``knowledge boundary''  and to assess if and to what extent the model predictions can be considered reliable. This property is orthogonal to the other uncertainty evaluation metrics and therefore needs to be separately evaluated.

The importance of calibration with respect to domain shifts has been highlighted in other contexts\cite{lakshminarayanan2017simple}, but its role in the chemical domain is even more prominent. Indeed, generalization power is a requirement in key applications such as drug discovery, and the intrinsic high variability of chemical space makes it challenging to fulfill this requirement.
Despite this  prominent role, the evaluation of out-of-domain uncertainty performance in the chemistry field appears to be absent\cite{Zhang2019} or very limited\cite{Ryu2018}, thus demanding a more extensive analysis. 

To achieve this goal, we employ the recently introduced \emph{scaffold splitting} technique\cite{wu2018moleculenet,Yang2019}. Molecules are split into bins based on their Murcko scaffold, with each  bin belonging to only one among training, validation and test set\cite{Yang2019}. 
Scaffold splitting has been successfully used to evaluate models under the more realistic assumption of significantly diverse training and testing distributions, thus overcoming the traditional random splitting. It has been demonstrated to be more challenging for a model and capable of simulating the chronological split which characterizes real scenarios of molecular property prediction\cite{Yang2019}.
To the best of our knowledge, scaffold splitting has never been used to evaluate out-of-domain uncertainty estimation procedures before.

More specifically, we are interested in \emph{re-evaluating} all the already introduced metrics --- AUCO, AUCE, etc. --- also in the \emph{out-of-domain} context obtained through scaffold-splitting. We will pay particular attention to \emph{out-of-domain calibration}, since it can measure to what extent  a model knows what it does not know.
We are interested in quantifying domain shift uncertainty performance, i.e., the ratio between  in-domain and  out-of-domain  metrics, also in relation to domain shift error (the ratio between in-domain and out-of-domain error) to assess if and to what extent error generalization and uncertainty generalization 
are characterized by the same behavior.

\section{Experiments}

We first describe the target dataset, followed by a description of the experimental procedure.

\subsection{Data} \label{ssec:data}
The formation enthalpies of 131,722 stable organic molecules composed of C, H, O, and N atoms were used to train and test the model. These reference data were derived from the QM9 dataset, which was calculated at the B3LYP/6-31G(2df,p) level of theory with the rigid rotor-harmonic oscillator approximation (RRHO).\cite{Ramakrishnan2014} As discussed in previous work, these calculated enthalpies are themselves associated with significant errors, primarily due to weaknesses of B3LYP such as the absence of long-range dispersion interaction but also the lack of rotor or conformer corrections in the calculations.\cite{cohen_challenges_2012,simm_systematic_2016,proppe_uncertainty_2017,li_thermodynamics_2016} We note that it is possible to use a small amount of high-accuracy coupled cluster training data via a transfer learning approach to minimize the influence of DFT errors. Interested readers are referred to the recent work of Grambow et al.\cite{grambow_accurate_2019}. In this work, we use the QM9 data as is without any attempt to correct its errors in order to investigate the effects of aleatoric uncertainties. The enthalpy values used for training and testing can be found in the Supporting Information. 

We used a 80:10:10 split for training, validation, and test sets, both in the in-domain and out-of-domain settings. Random splitting has been used for in-domain analysis, while, as previously discussed, scaffold splitting has been used for out-of-domain analysis. In both cases, the same split has been employed to test all the methods.

\subsection{Experimental Procedure} \label{ssec:exp_procedure}

We evaluated the uncertainty estimation techniques previously reviewed using the methods previously introduced. Other than including diagrams, we evaluated the considered methods quantitatively, as follows:
\begin{itemize}
    \item For ranking-based evaluation we use the Area Under the Confidence-Oracle error (\textit{AUCO}) as a measure of total discrepancy with respect to the best possible ranking, the \textit{Error Drop} as a measure of total error reduction for high-confident predictions  and the \textit{Decrease Ratio} to assess the monotonicity of confidence curves.
    \item For confidence-based calibration we use the Area  Under  the  Calibration  Error Curve (\textit{AUCE}) as a measure of total discrepancy with respect to perfect calibration and the Maximum Calibration Error (\textit{MCE}) to account for the worst-case scenario\footnote{We did not use Expected Calibration Error (ECE) in our tests because it does not add significant information to AUCE for confidence-based calibration.}.
    \item For error-based calibration we use the Expected Normalized Calibration Error (\textit{ENCE}) as a measure of the (normalized) total discrepancy with respect to perfect calibration.
    \item For dispersion evaluation we use the coefficient of variation $c_v$.
    \item For domain-shift performance we evaluated and compared all the above metrics also in an \emph{out-of-domain} setting obtained using scaffold-splitting, as previously detailed.
\end{itemize}

We focused on the evaluation of complete and scalable uncertainty frameworks, therefore we compared MC-Dropout (with Concrete Dropout, as previously discussed), ensembling and bootstrapping.
As previously mentioned, these approaches have been designed to model NN-weight uncertainties, therefore they are directly related to epistemic uncertainty estimation. However, they have been used and described in the literature in conjunction with aleatoric uncertainty estimation to form complete frameworks \cite{gal2016dropout,lakshminarayanan2017simple}, and this is the way we tested them in this work.
In addition to evaluating total uncertainty, we have also separately evaluated aleatoric and epistemic uncertainty for each methodology. All the different methods use the same aleatoric approximation scheme but the way epistemic uncertainty is modeled affects also aleatoric uncertainty results, thus resulting in different outputs (ref. Eq. \eqref{eq:mean_variance}). This also allows  drawing conclusions about aleatoric uncertainty which do not depend on the uncertainty model used for the NN-weights.

\subsubsection{Implementation and experimental setting}
We implemented the tested uncertainty estimation methods starting from the base model  made available in \citet{Yang2019}, based on the PyTorch framework. 

We performed hyperparameter optimization using the \texttt{hyperopt} package\footnote{\url{https://github.com/hyperopt/hyperopt}} on the base model and we used the same hyperparameters for all the uncertainty methods tested. The hyperparameters are: depth size for the convolutional layer $= 6$, depth size for the fully connected layer $= 2$, hidden size $= 1000$. 
The number of instances is 15 for ensembling and bootstrapping and 150 for MC-Dropout\footnote{MC-Dropout employs weight sharing between different instances and  it does not require a separate training for each one, allowing the usage of more instances in practice. Therefore, this difference in the number of instances reflects  realistic condition of use.}.

All the results obtained are inevitably a function of the number of instances used, since the approximation performance of all the tested methods depends on it. The number of instances chosen for the experiments is in line with what has been described in the literature; additionally, preliminary experiments varying the number of instances did not report significant variations in the outcomes, except for an asymptotically smaller general improvement in all the metrics for all the tested methods. 

\section{Results} \label{sec:results}
We first detail error performance for the considered models. Next, we present results for uncertainty estimation evaluation.

\subsection{Error} \label{ssec:results_error}

Table \ref{tab:error} lists the mean absolute error (MAE) for the considered models both in the \emph{in-domain} and \emph{out-of-domain} settings.

\begin{table}
  \caption{Mean absolute error (MAE) on the test dataset (kcal/mol). Results are shown for the base model without any uncertainty estimation and for the model extended with each of the evaluated uncertainty estimation methods. Both in-domain and out-of-domain performance are reported for each case.}
  \label{tab:error}
  \begin{tabular}{lll}
    \hline
      & In domain &  Out domain \\
    \hline
    Base model   & 1.04 &  1.77 \\
    MC-Dropout & 0.97 & 1.49 \\
    Ensembling  & 0.74  & 1.21 \\
    Bootstrapping & 0.89 & 1.43 \\
    \hline
  \end{tabular}
\end{table}

The baseline is the \texttt{chemprop} model\cite{Yang2019} without any uncertainty estimation. We notice how extending it to include uncertainty always leads to reductions in MAE, regardless of the approximation method used (MC-Dropout, ensembling and bootstrapping).  These improvements, often underestimated, are due to both aleatoric and epistemic estimation in the model. Indeed, modelling aleatoric uncertainty implicitly reduces the impact of noisy training samples, thus improving predictive performance. Modelling epistemic uncertainty allows averaging multiple weight configurations, avoiding overfitting, and overconfident estimations, with a positive impact on predictions. These two contributions can independently reduce the overall MAE but act synergistically when both are modeled.
We can notice that, independently from the model, the reduction in out-of-domain error is higher than in-domain error.

The analysis of improvements in MAE is not the main goal of the present paper, but its assessment is useful for the following discussion and should be kept in consideration as an important by-product of Bayesian uncertainty modelling.

\subsection{Uncertainty estimation} \label{ssec:results_uncertainty_estimation}

\subsubsection{Ranking-based evaluation}

\begin{figure}
    \centering
    \includegraphics[width=1.0\columnwidth]{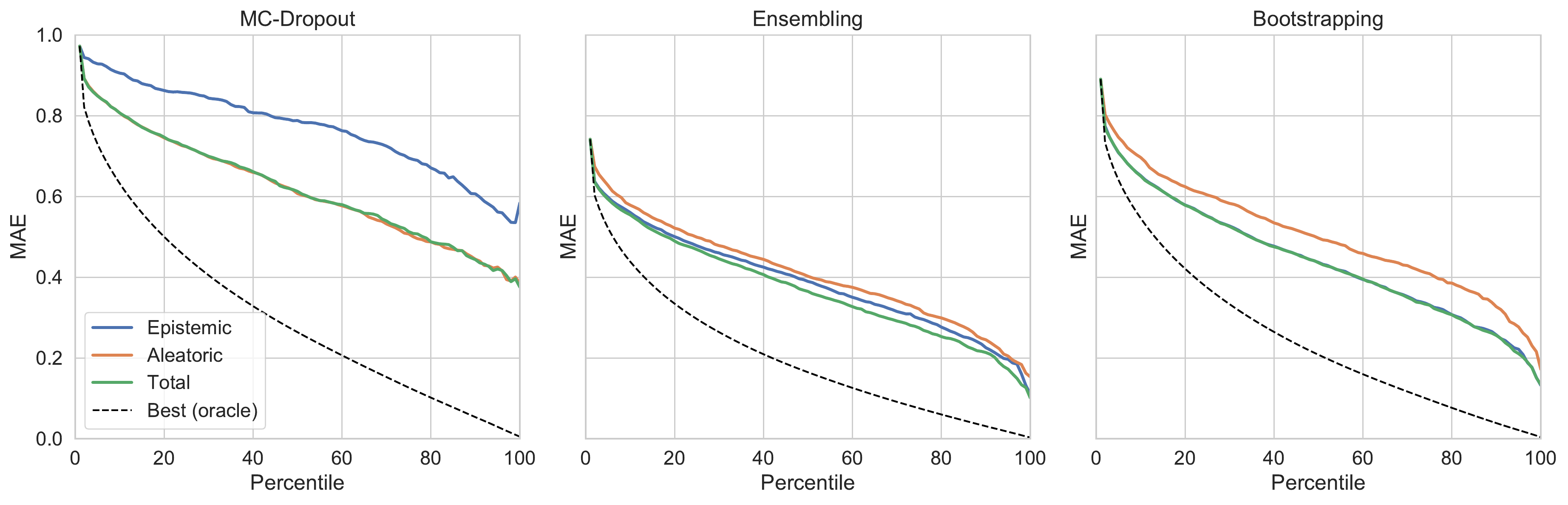}
    \caption{Confidence curves for the different methods (in-domain). MAE in kcal/mol.}
    \label{fig:confidence}
\end{figure}

\begin{figure}
    \centering
    \includegraphics[width=0.55\columnwidth]{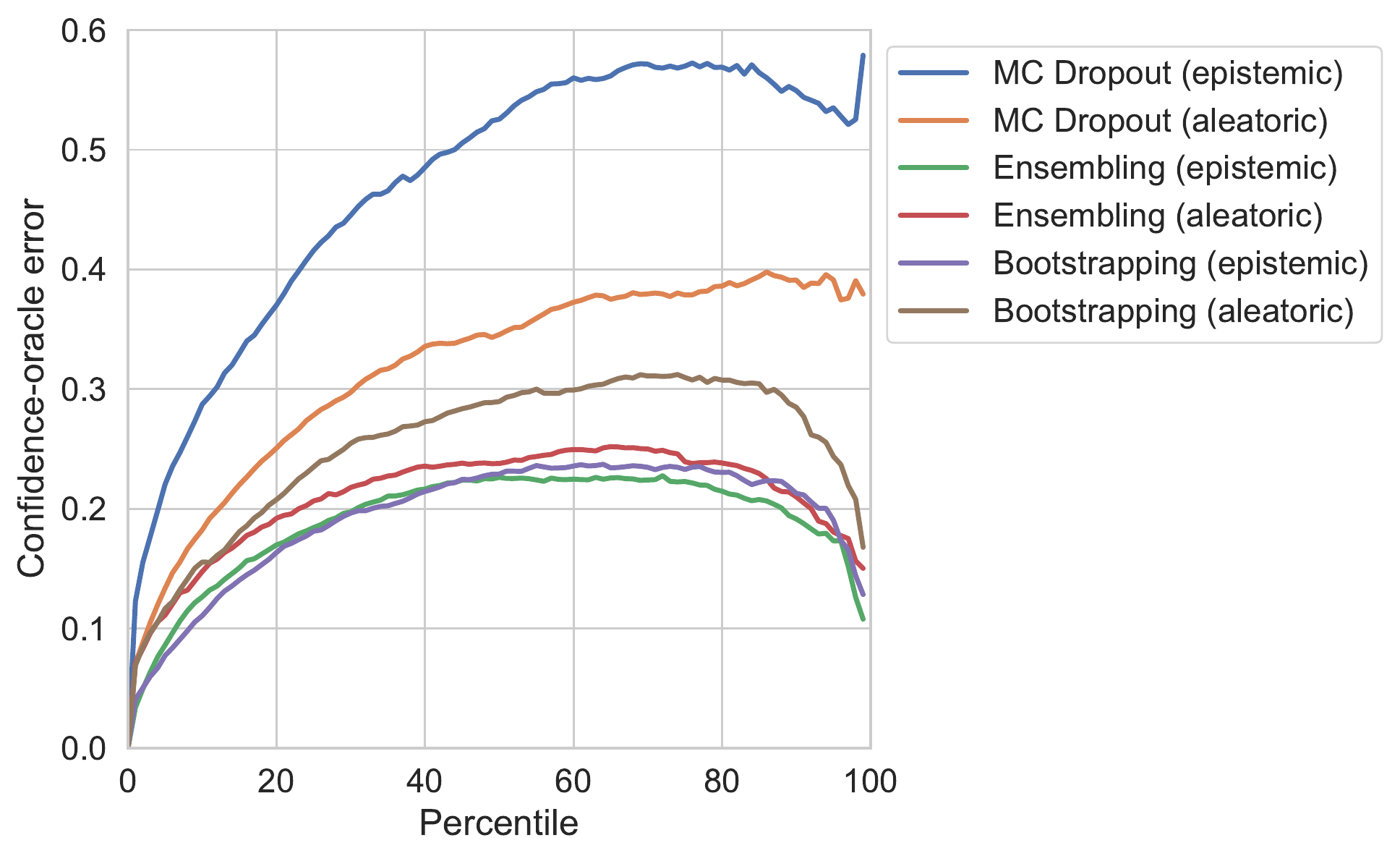}
    \caption{Confidence-Oracle errors for the different methods (in-domain). Error in kcal/mol.}
    \label{fig:confidence-oracle}
\end{figure}

The confidence curves for the different methods and the related Confidence-Oracle errors are shown in Fig. \ref{fig:confidence} and Fig. \ref{fig:confidence-oracle}, respectively.
The derived AUCO and Decrease Ratio metrics for each case are reported in the first two lines of Table \ref{tab:summary}.

We can observe that all the curves are mostly decreasing, therefore each method can establish a qualitatively meaningful ranking of the predictions by their uncertainty. However, as also highlighted by the Decrease Ratio, MC-Dropout does not lead to perfectly non-increasing curves, especially for epistemic uncertainty and at high percentiles.

In absolute terms, ensembling allows reaching the lowest MAE in the highest percentiles in both the components and the total uncertainty. Interestingly, the epistemic uncertainty estimated by bootstrapping allows reaching a MAE comparable to ensembling in the highest percentiles (0.21 versus 0.19 kcal/mol in the top 5\%), even if the initial MAE on the whole dataset is significantly worse (0.89 versus 0.74 kcal/mol). This is quantitatively measured by a higher or similar error drop of bootstrapping, despite the overall higher MAE.

To compare the relative performance of the different approaches we need to consider the Confidence-Oracle errors and the AUCO. Globally, ensembling results in the lowest errors, even if the epistemic uncertainty estimated by bootstrapping leads to comparable performance. In contrast, the aleatoric component of bootstrapping leads to a significantly worse performance than ensembling. MC-Dropout results in larger errors with respect to the other considered approaches, in particular for epistemic uncertainty. 

The total uncertainty does not always result in a lower (i.e., better) AUCO than the two separate contributions.
While this is true  for ensembling, it is not true in the other cases.
In general, in ranking-based evaluation, if $\sigma_a \gg \sigma_e$ or vice-versa, the total uncertainty curve will approximate the dominant contribution.
Anyway, as we can observe, in these cases the total uncertainty appears to approximate the best performing one in terms of AUCO. 

\subsubsection{Calibration and dispersion}

\paragraph{Confidence-based calibration}

\begin{figure}
    \centering
    \includegraphics[width=1.0\columnwidth]{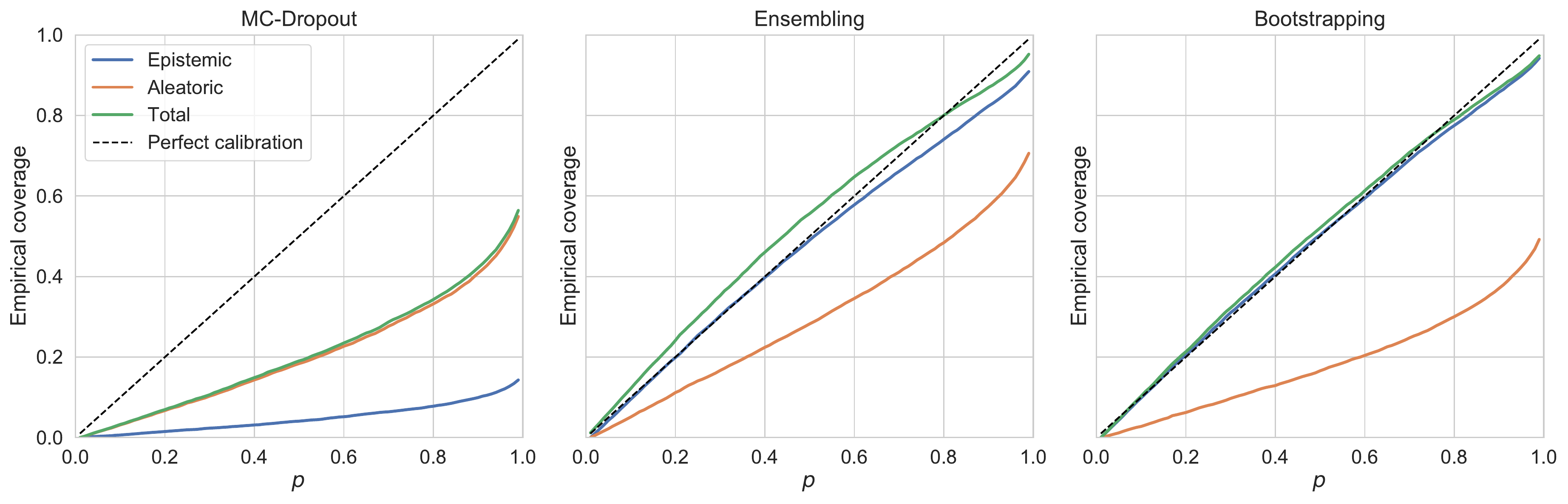}
    \caption{Confidence-based calibration for the different methods (in-domain).\\
    The \emph{empirical coverage} is reported for each symmetric confidence interval of probability $p$ defined by the uncertainty. The empirical coverage is the fraction of times the true value actually falls in a confidence interval.}
    \label{fig:calibration-confidence}
\end{figure}

The confidence-based calibration plots are shown in Fig. \ref{fig:calibration-confidence}. The derived AUCE and MCE metrics are reported in lines three and four of Table \ref{tab:summary}, respectively.

Results for epistemic uncertainty vary. Ensembling is characterized by calibrated empirical coverages in the low probability range ($ p<50\%$), but increasingly  underestimated coverages in the high probability range. Bootstrapping has a similar pattern but is better calibrated overall, with a broader interval of calibrated empirical coverages ($p<75\%$) and less underestimated coverages for higher values.  
This is quantified by the AUCE, which captures the overall behavior and is halved for bootstrapping with respect to ensembling. MC-Dropout epistemic uncertainty is largely underestimated.

In general, aleatoric uncertainty  appears to be underestimated, independently from the underlying uncertainty model of the NN weights. The possible reasons for a miscalibrated aleatoric uncertainty are discussed in the last section.

Total uncertainty does not result in significant improvements to AUCE compared to considering epistemic uncertainty only in any of the cases, leading instead to slightly worse performance for ensembling and bootstrapping. By contrast, MCE is improved in those cases due to the combination of an underestimated aleatoric uncertainty and an overestimated epistemic uncertainty, which results in more stable curves. This also highlights the need of multiple metrics to quantify calibration.

\paragraph{Error-based calibration}

\begin{figure}
    \centering
    \includegraphics[width=0.7\columnwidth]{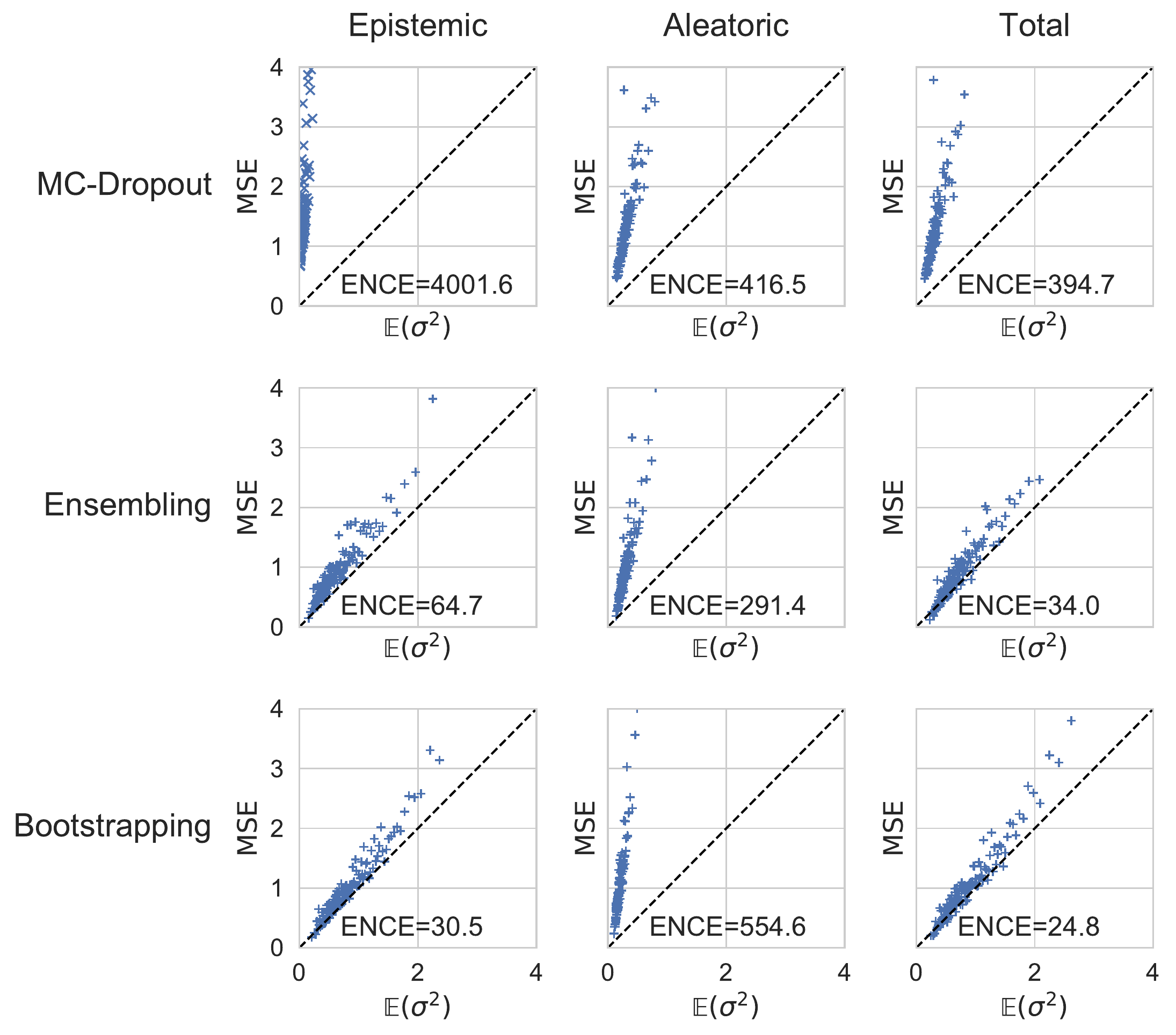}
    \caption{Error-based calibration for the different methods (in-domain). MSE in (kcal/mol)$^2$.}
    \label{fig:calibration-error}
\end{figure}

The error-based calibration plots are shown in Fig. \ref{fig:calibration-error}. The derived ENCE is also reported in line five  of Table \ref{tab:summary}.

These plots offer a complementary view of uncertainty performance with respect to the confidence-based plots already shown. Indeed, rather than considering all the predictions at the same time, each dot only represents a subset of predictions in direct relation with the average error.

Aleatoric uncertainty on its own significantly underestimates the error in all the cases. Epistemic uncertainty appears to be a better error approximator for ensembling and bootstrapping, with a lead of the latter ($64.7$ vs $30.5$ AUCE), but not for MC-Dropout.
Total uncertainty always reports a better AUCE than the two individual contributions.
Uncertainty tends to be underestimated in all of the considered cases. 

Compared to confidence-based calibration, this kind of plot is less stable, especially for high values of $\sigma$. This is due to i) the fact that the error is expected to be naturally higher as uncertainty increases (a property already taken into account in the ENCE computation) and ii) the fact that high uncertainty values are more sparse. 
Overall, error-based calibration confirms the main results of confidence-based calibration: bootstrapping estimates appear to be better calibrated and the total uncertainty is a better error approximator.

Interestingly, we notice that all the plots, independently from their distance to the diagonal line,  are characterized by strongly correlated patterns (correlation $\approx 0.90\sim0.93$ for ensembling and bootstrapping, $\approx 0.66\sim0.87$ for MC-Dropout).

\paragraph{Dispersion}

The dispersion coefficient is reported in the last line of Table \ref{tab:summary}. Results show no significant variations between the different methods, except for a slightly higher $c_v$ for  MC-Dropout epistemic estimates. In general, epistemic uncertainty appears to be more disperse than aleatoric uncertainty for all the considered methods.

\begin{table}
  \caption{Summary metrics for the different methods (in-domain).}
  \label{tab:summary}
  \begin{tabular}{c|ccc|ccc|ccc}
    \hline
    & \multicolumn{3}{c|}{MC-Dropout} & \multicolumn{3}{c|}{Ensembling} & \multicolumn{3}{c}{Bootstrapping} \\
      & Epi. &  Ale. & Tot. & Epi. &  Ale. & Tot. & Epi. &  Ale. & Tot. \\
    \hline
    AUCO   & 46.72 & 31.55  & 31.72 &  18.79 & 20.83 & 17.03 & 19.18 & 25.08 & 19.05 \\
    Error drop & 1.67 & 2.55  & 2.62 &  6.72 & 4.93 & 7.40 & 6.85 & 5.23 & 6.85 \\
    Decr. Ratio & 0.95 &  0.98 & 0.96 & 1.0 & 0.99 & 1.0 & 1.0 & 1.0 & 1.0 \\
    AUCE & 44.79 &  29.44 & 28.74 & 2.62 & 19.90 & 3.62 & 1.36 & 31.31 & 1.69 \\
    MCE   & 0.85 &  0.50 & 0.48 & 0.087 & 0.33 & 0.061 & 0.051 & 0.53 & 0.044 \\
    ENCE   & 4001.6 &  416.5 & 394.7 & 64.7 & 291.4 & 34.0 & 30.5 & 554.6 & 24.8\\
    $c_v$   & 0.97 & 0.50 & 0.49 & 0.74 & 0.51 & 0.67 & 0.74 & 0.45 & 0.71\\
    \hline
  \end{tabular}
\end{table}

\subsubsection{Out-of-domain uncertainty}
The same plots already discussed for random splitting are shown for the out-of-domain case. The derived metrics are summarized in Table \ref{tab:summary_scaffold}. In the following, the main differences with respect to random splitting are highlighted.

\begin{figure}
    \centering
    \includegraphics[width=1.0\columnwidth]{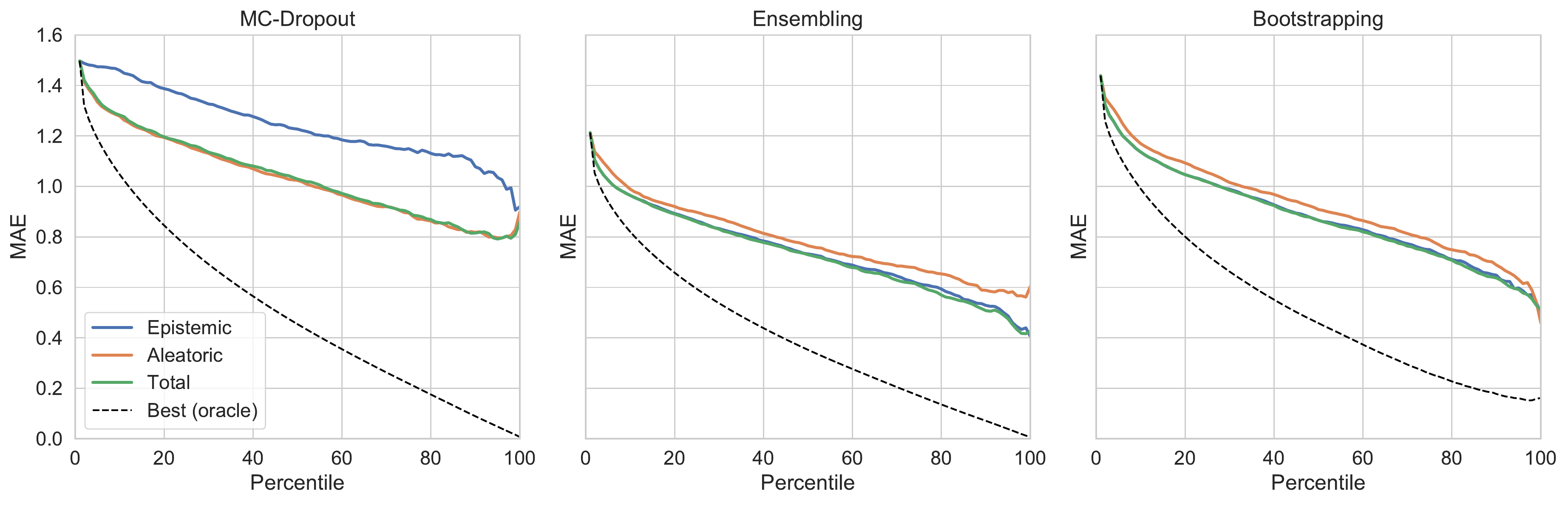}
    \caption{Confidence curves for the different methods (out-of-domain). MAE in kcal/mol.}
    \label{fig:confidence_scaffold}
\end{figure}

\begin{figure}
    \centering
    \includegraphics[width=0.55\columnwidth]{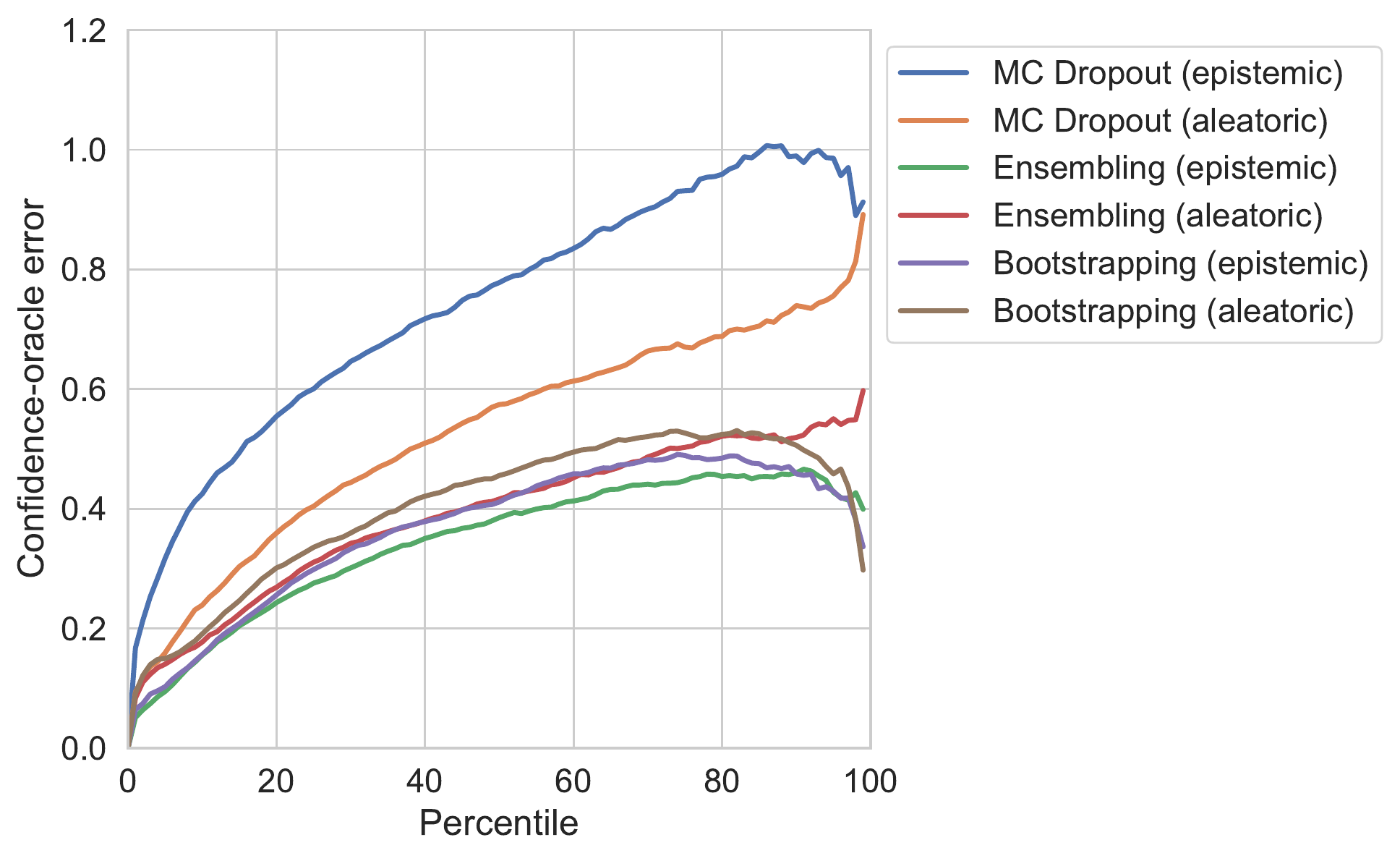}
    \caption{Confidence-Oracle errors for the different methods (out-of-domain). Error in kcal/mol.}
    \label{fig:confidence-oracle_scaffold}
\end{figure}

Confidence curves and Confidence-Oracle errors for the out-of-domain case are reported in Fig. \ref{fig:confidence_scaffold} and Fig. \ref{fig:confidence-oracle_scaffold}, respectively.
In absolute terms, as expected all the related out-of-domain indices (AUCO, error drop and decrease ratio) have deteriorated with respect to in-domain indices for all the considered methods. The relative performance of  MC-Dropout with respect to ensembling and bootstrapping are comparable, with  these last two outperforming the first.
The relative comparison between ensembling and bootstrapping results in qualitatively similar trends but quantitative differences which turn out to be strongly reduced.
Ensembling has the lowest AUCO for both epistemic and aleatoric uncertainty, bootstrapping has comparably low scores and it also has comparably or higher error drops. The results for these two methods turn out to be more similar than in the in-domain setting.
In general, the ranking-based evaluation in the out-of-domain setting does not highlight drastic changes other than an expected worsening of all the indices for all the methods.

\begin{figure}
    \centering
    \includegraphics[width=1.0\columnwidth]{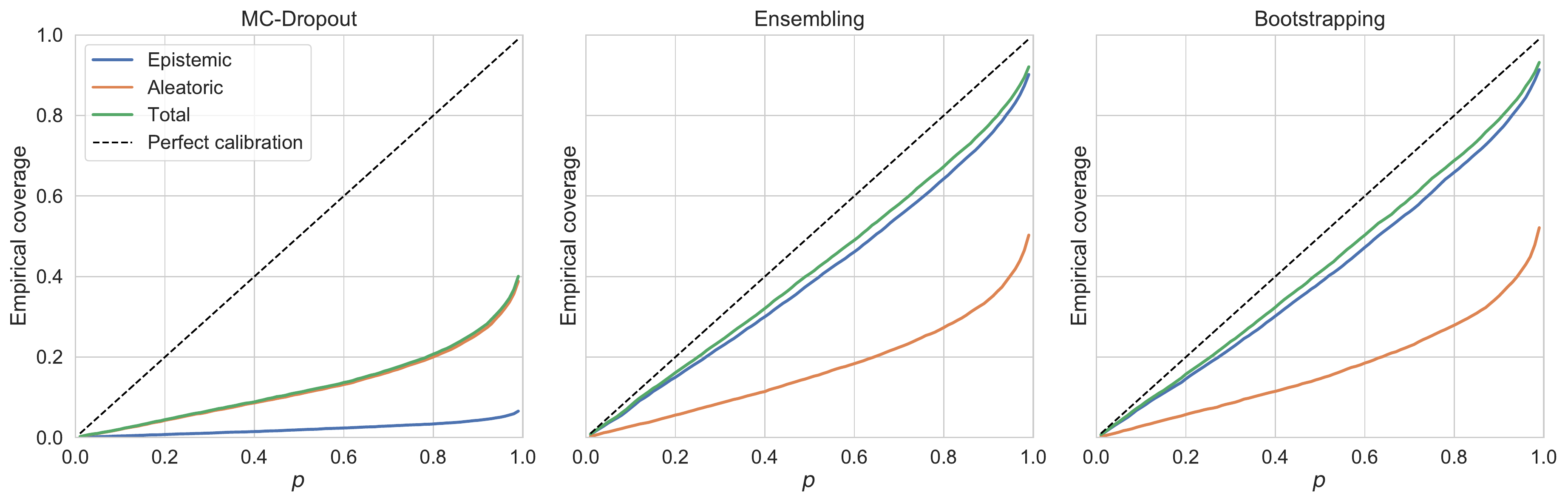}
    \caption{Confidence-based calibration for the different methods (out-of-domain). \\
    The \emph{empirical coverage} is reported for each symmetric confidence interval of probability $p$ defined by the uncertainty. The empirical coverage is the fraction of times the true value actually falls in a confidence interval.}
    \label{fig:calibration-confidence_scaffold}
\end{figure}

\begin{figure}
    \centering
    \includegraphics[width=0.7\columnwidth]{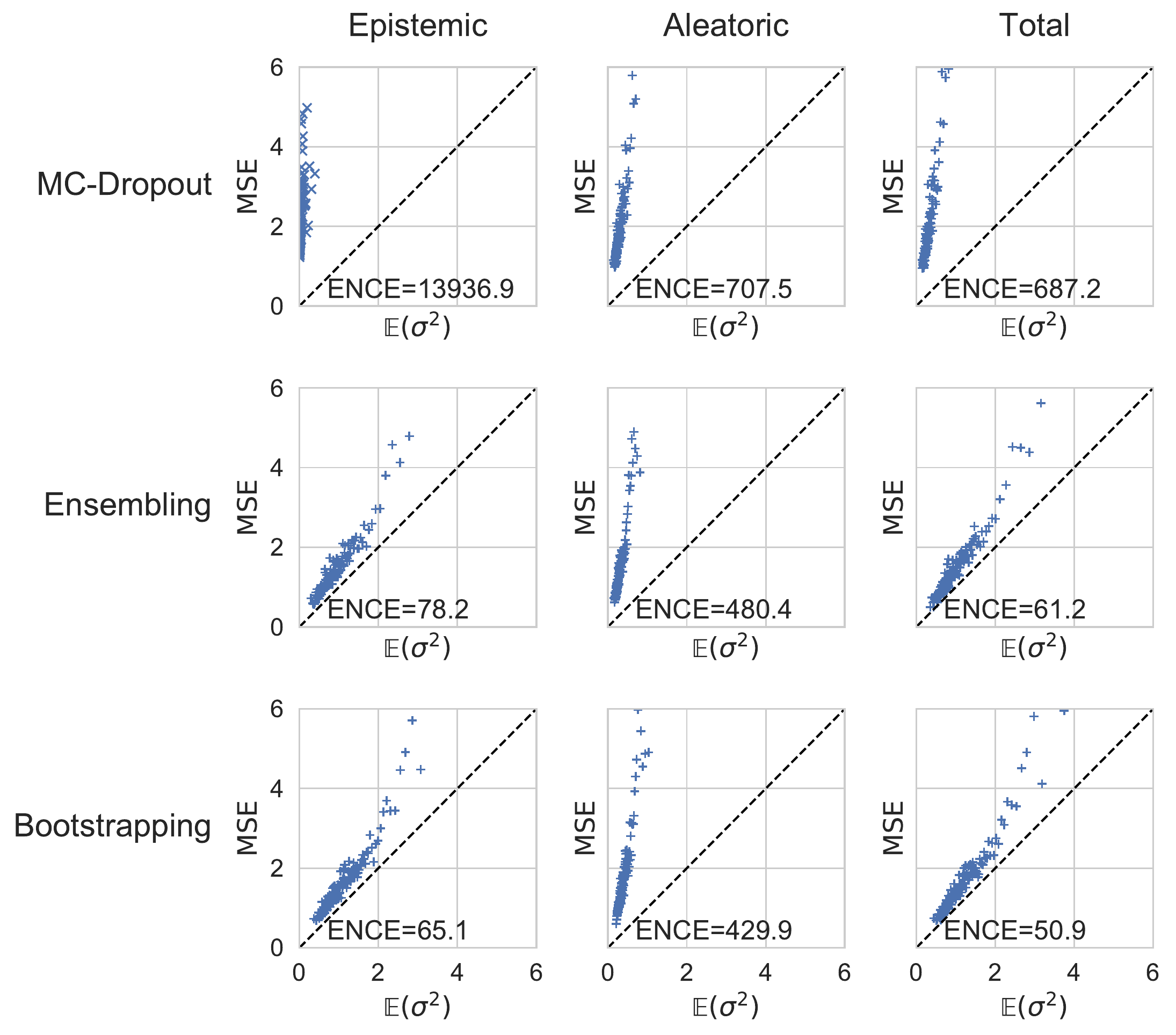}
    \caption{Error-based calibration for the different methods (out-of-domain). MSE in (kcal/mol)$^2$.}
    \label{fig:calibration-error_scaffold}
\end{figure}

The calibration-confidence analysis (Fig. \ref{fig:calibration-confidence_scaffold} and Fig. \ref{fig:calibration-error_scaffold}) highlights a drastic change with respect to in-domain results for epistemic estimates using ensembling and bootstrapping. In particular, while in-domain empirical coverages tend to be calibrated or slightly overestimated, except for high $p$, out-of-domain empirical coverages tend to be always underestimated.
This means that, on average, uncertainty estimates in an out-of-domain setting are lower than they should, while in-domain uncertainty estimates appear to be more calibrated or slightly higher than they should. 
Aleatoric estimates are less affected than epistemic ones  in terms of AUCE and MCE for all the considered methods. Calibration-error analysis confirms the underestimation trend of out-of-domain epistemic estimates,  particulary affecting high-error predictions. The impact of out-of-domain uncertainty underestimation is further discussed in the next section.

Overall, bootstrapping has a slight  advantage over ensembling in terms of AUCE, MCE and ENCE driven both by  better epistemic uncertainty estimates (even if the magnitude of the difference is less than in-domain) and also better aleatoric uncertainty estimates (in contrast to in-domain results). This highlights another difference with respect to in-domain analysis, that is further discussed in the next section.

An additional difference pointed out by calibration analysis concerns the total uncertainty. While in-domain total uncertainty turns out to be similar or slightly worse than the two individual components, out-of-domain total calibration appears to be better than the two individual components for all the considered metrics.

In terms of dispersion, we observe a global increase for all the methods and uncertainty types.

\begin{table}
  \caption{Summary metrics for the different methods (out-of-domain).}
  \label{tab:summary_scaffold}
  \begin{tabular}{c|ccc|ccc|ccc}
    \hline
    & \multicolumn{3}{c|}{MC-Dropout} & \multicolumn{3}{c|}{Ensembling} & \multicolumn{3}{c}{Bootstrapping} \\
      & Epi. &  Ale. & Tot. & Epi. &  Ale. & Tot. & Epi. &  Ale. & Tot. \\
    \hline
    AUCO   & 73.35 & 52.29  & 52.93 &  34.12 & 38.68 & 33.38 & 36.27 & 40.05 & 35.81 \\
    Error drop   & 1.64 & 1.67  & 1.75 &  3.02 & 2.02 & 2.88 & 2.91 & 3.18 & 2.86 \\
    Decr. Ratio & 0.86 & 0.90 & 0.91 & 0.99 & 0.95 & 0.98 & 0.98 & 0.99 & 1.0 \\
    AUCE & 47.36 &  37.13 & 36.70 & 10.18 & 32.81 & 8.10 & 9.62 & 32.57 & 7.50 \\
    MCE   & 0.92 &  0.65 & 0.64 & 0.16 & 0.56 & 0.13 & 0.14 & 0.55 & 0.11 \\
    ENCE   & 13936.2 &  707.5 & 687.2 & 78.2 & 480.4 & 61.2 & 65.1 & 429.9 & 50.9\\
    $c_v$   & 1.63 &  0.76 & 0.75 & 0.65 & 0.84 & 0.65 & 0.81 & 0.71 & 0.79\\
    \hline
  \end{tabular}
\end{table}

\section{Discussion} \label{sec:discussion}

The goal of this section is to analyze and discuss the results presented in previous section, focusing on conclusions that can be drawn by comparing and integrating outcomes related to different uncertainty models and evaluation metrics.

Results show that ensembling and bootstrapping consistently outperform MC-Dropout both in the in-domain and out-of-domain scenarios for all the considered metrics. This is in line with results already presented for image classification/regression\cite{lakshminarayanan2017simple,beluch2018power} and optical flow estimation\cite{ilg2018uncertainty,gustafsson2019evaluating}, confirming this trend also for GCNN-based  molecular property prediction. In contrast to previous comparisons, that used the  ``base'' version of MC-Dropout\cite{lakshminarayanan2017simple,ilg2018uncertainty,gustafsson2019evaluating}, we employed Concrete MC-Dropout that was independently proven superior to standard MC-Dropout\cite{gal2017concrete,mukhoti2018evaluating} but has not been directly compared to ensembling and bootstrapping before.

The comparison between ensembling and bootstrapping requires a deeper analysis and raises multiple interesting observations. On the one side, ensembling has an advantage for total MAE, AUCO and aleatoric calibration, especially in the in-domain setting. On the other, bootstrapping often leads to higher error drops (i.e. it allows reducing the MAE more in proportion when we consider small percentages of high-confidence predictions), has an advantage for better epistemic calibration in the in-domain setting and is characterized by an overall better calibration in the out-of-domain setting.
This behavior can be explained by considering the effects of substituting each training dataset with a bootstrap sample. Each network only sees a fraction of  the starting training dataset, thus increasing individual and ensembled MAE. Since aleatoric uncertainty is estimated from data, it follows a trend similar to MAE and it degrades. However, bootstrapping promotes diversity in ensembled models, which is key for epistemic uncertainty estimation, thus improving its calibration. We can argue that as training size increases --- as long as the target molecular space is kept unchanged --- bootstrapping becomes more advantageous, because each bootstrap sample becomes a better approximator of the underlying distribution, thus avoiding losses in MAE and aleatoric calibration in each single instance and in the ensembled model, but keeping an advantage as for epistemic calibration.
Moreover, as we have observed, bootstrapping becomes globally more calibrated than ensembling in the out-of-domain setting. This can be explained by a gain of generalization power given by  the additional diversity of bootstrapping. Interestingly, this generalization power  especially translates in calibration performance, and only to a lesser extent in ranking-based indices and total MAE, which turn out to be relatively improved in the out-of-domain setting with respect to ensembling, but not better than the latter in absolute terms.
Dispersion analysis allows checking that improvements in calibration are \emph{not} the result of losses in uncertainty heterogeneity. 

In previous studies for CNN-based image regression/classification, bootstrapping did not report significant improvements over ensembling \cite{lakshminarayanan2017simple}. We can speculate that this difference is due to i) the peculiarities of the chemical space, characterized by a larger intrinsic variability that can be exploited by bootstrapping, and ii) by variations in the training size, as previously discussed.
Results obtained for bootstrapping  justify its recent use in active learning methodologies  for molecular property prediction\cite{Li2019}, where model uncertainty (epistemic uncertainty) and generalization power are required. 

Even if the methods investigated in this work jointly model aleatoric and epistemic uncertainties, their separate evaluation carried out in the previous section allows directly comparing the two. 
Both appear to be effective for ranking-based evaluation, with a potential complementary improvement  of total uncertainty. From a calibration point of view, good performance has been reached using epistemic uncertainty alone, while aleatoric uncertainty individually turns out to always be largely underestimated, even if it is characterized by a high correlation with error. In any case, total uncertainty is as calibrated as the individual components, and even more calibrated in the out-of-domain setting.
We can explain this behavior of calibration as follows.

Aleatoric uncertainty should correlate with the noise in the observed variable, while epistemic uncertainty with the error in the trained function. However, the only observable error (MSE) includes both these contributions. 
Therefore, we can speculate that in this specific case epistemic uncertainty appears to be more calibrated than aleatoric uncertainty individually because the total error is primarily due to the model's approximating function rather than the noise in the data. In other contexts, the individual contributions to total error could vary, and the situation could be reversed, but MSE should always be better approximated by total uncertainty. Evaluating the individual contributions can be helpful in pinpointing their relative importance in different settings.
Moreover, even if MSE is better approximated by total uncertainty, applications could require taking into account only one of the two components for its specific meaning or to maximize some specific metric.
This kind of analysis is not the main goal of this work and deserves further investigation.

Domain shift analysis is characterized by mixed results. On the one side, ranking-based performance does not appear to be particularly affected by out-of-domain molecules: the AUCO decreases proportionally to the (inevitable) decrease in total MAE, while the error drop is even larger than in the in-domain setting. On the other side, calibration performance drastically changes and out-of-domain calibration appears to be \emph{consistently underestimated}. The latter result is in line with what has been recently observed in \citet{Li2019}, but the analysis carried out in this work has allowed the quantification of this behavior and its confirmation in a more general setting with multiple uncertainty methods being employed.
As the model is tested on molecules different with respect to those seen during training, the error increases without the uncertainty being able to totally capture this rise, thus leading to lower than expected estimates in this case. 
Out-of-domain uncertainty calibration should be a major focus of future development in uncertainty estimation methodologies for molecular property prediction.

Up to now, we mainly compared uncertainty models. However, the obtained results also allow for the comparison of different \emph{evaluation methods} in terms of what they capture about uncertainty to discuss if and to what degree they are all necessary and complementary.
Taking into consideration calibration allows identifying several patterns that do not emerge from confidence curves only, such as the discrepancy in ensembling epistemic and aleatoric uncertainties  or some differences between ensembling and bootstrapping, thus highlighting its important role in comparisons. By contrast, even recent work that seeks to obtain ``uncertainty-calibrated prediction of molecular properties''\cite{Zhang2019} do not take into consideration calibration evaluation in the results.
The discrepancy between results obtained based on the two different definitions of calibration is more subtle. Qualitatively, the main conclusions derived by confidence-based calibration, such as the largely underestimated aleatoric uncertainty in all the experiments, are also reflected in error-based calibration. Quantitatively, the ratios of the indices obtained through these two methods do not always overlap, but they always rank models in the same order. Based on the obtained results, it is not possible to state if and when quantitative indices based on one of the two definitions outperform the other. 
The results obtained for these two different definitions of calibration also confirm their previous comparative discussion. In particular, even if error-based calibration directly relates error and uncertainty according to the definition, the inherent non-uniformity of uncertainty estimates makes it difficult to obtain reliable statistics in some uncertainty ranges (high uncertainty ranges in our experiments), with less stable results. This also prevents assessing if the error in these ranges is due to uncertainty estimates themselves or to insufficient data for computing reliable statistics. 
Therefore, we can conclude that the choice between these two evaluation techniques depends on the context. If the dataset is large enough to enable meaningful estimates for all the bins, error-based calibration should be preferred because it allows for a more direct comparison and it avoids issues when re-calibration techniques are employed \cite{levi2019evaluating}. Instead, if the uncertainty distribution is highly skewed and few samples are available in some ranges, as it turns out in our experiments, confidence-based calibration can overcome this and results in less noisy plots. 

\section{Conclusion and Future Work}

In this paper we compared three state-of-the-art  approaches for uncertainty estimation in neural networks in the context of GCNNs for molecular property prediction: MC-Dropout with Concrete Dropout, ensembling, and bootstrapping. We selected those approximate Bayesian inference techniques satisfying some specific application-oriented criteria: scalability, lack of hyper-parameters, and independence from the underlying network architecture. 
These techniques have been first reviewed in a unified framework that separates aleatoric and epistemic uncertainty, also in the light of recent interpretations given to ensembling, and then experimentally compared on the QM9 dataset based on a set of introduced criteria. Those criteria have been selected to evaluate uncertainty from different perspectives: based on its ability to define a ranking of most confident predictions, based on uncertainty calibration (two different recent definitions for regression have been employed), based on dispersion that measures estimated heterogeneity, and based on robustness to domain shift in the test set with respect to the training set, with scaffold splitting being employed. 

The obtained results lead to multiple interesting conclusions. First of all, ensembling and bootstrapping appear to consistently outperform MC-Dropout, confirming the results recently presented for other domains and different network types also for GCNN-based molecular property prediction.
The comparison between ensembling and bootstrapping leads to more mixed results.
Even though ensembling is better with respect to most of the considered metrics, including overall MAE, bootstrapping appears to outperform ensembling for others, notably epistemic uncertainty calibration and overall out-of-domain calibration. This is not in line with what has been previously described in the context of image regression/classification, highlighting an interesting property of the chemical space and/or the chemical dataset analyzed.
Furthermore, the results presented have led to a better understanding about the role of aleatoric/epistemic uncertainty with an interesting method based on calibration plots to pinpoint the relative contribution of the two kinds of uncertainty to the total error.

The latter is one of the directions that should be further investigated in the future, with a deeper analysis of the uncertainty components, also in relation to the specific features of the datasets. In addition, taking into consideration how approximate methods interfere with their independent calculation would be of crucial importance in applications.
Another important direction concerns the improvement of uncertainty estimation methods. To accomplish this, a promising direction --- especially for epistemic and out-of-domain uncertainty --- is represented by the increase of diversity in the ensembled networks. This might not be the result of diversity in the data, as in bootstrapping, but instead come from the model itself\cite{lee2015m,pearce2018bayesian}. Balancing diversity, training data size and number of hyper-parameters appears to be a challenging tradeoff.
One of the main limitations of all the uncertainty estimation methods  is out-of-domain uncertainty calibration, and overcoming this weakness should be a major goal of future developments in uncertainty-aware molecular property prediction.

\bibliography{main}

\providecommand{\latin}[1]{#1}
\makeatletter
\providecommand{\doi}
  {\begingroup\let\do\@makeother\dospecials
  \catcode`\{=1 \catcode`\}=2 \doi@aux}
\providecommand{\doi@aux}[1]{\endgroup\texttt{#1}}
\makeatother
\providecommand*\mcitethebibliography{\thebibliography}
\csname @ifundefined\endcsname{endmcitethebibliography}
  {\let\endmcitethebibliography\endthebibliography}{}
\begin{mcitethebibliography}{56}
\providecommand*\natexlab[1]{#1}
\providecommand*\mciteSetBstSublistMode[1]{}
\providecommand*\mciteSetBstMaxWidthForm[2]{}
\providecommand*\mciteBstWouldAddEndPuncttrue
  {\def\EndOfBibitem{\unskip.}}
\providecommand*\mciteBstWouldAddEndPunctfalse
  {\let\EndOfBibitem\relax}
\providecommand*\mciteSetBstMidEndSepPunct[3]{}
\providecommand*\mciteSetBstSublistLabelBeginEnd[3]{}
\providecommand*\EndOfBibitem{}
\mciteSetBstSublistMode{f}
\mciteSetBstMaxWidthForm{subitem}{(\alph{mcitesubitemcount})}
\mciteSetBstSublistLabelBeginEnd
  {\mcitemaxwidthsubitemform\space}
  {\relax}
  {\relax}

\bibitem[Yang \latin{et~al.}(2019)Yang, Swanson, Jin, Coley, Eiden, Gao,
  Guzman-Perez, Hopper, Kelley, Mathea, Palmer, Settels, Jaakkola, Jensen, and
  Barzilay]{Yang2019}
Yang,~K.; Swanson,~K.; Jin,~W.; Coley,~C.; Eiden,~P.; Gao,~H.;
  Guzman-Perez,~A.; Hopper,~T.; Kelley,~B.; Mathea,~M.; Palmer,~A.;
  Settels,~V.; Jaakkola,~T.; Jensen,~K.; Barzilay,~R. Analyzing Learned
  Molecular Representations for Property Prediction. \emph{Journal of Chemical
  Information and Modeling} \textbf{2019}, \emph{59}, 3370--3388\relax
\mciteBstWouldAddEndPuncttrue
\mciteSetBstMidEndSepPunct{\mcitedefaultmidpunct}
{\mcitedefaultendpunct}{\mcitedefaultseppunct}\relax
\EndOfBibitem
\bibitem[Wu \latin{et~al.}(2018)Wu, Ramsundar, Feinberg, Gomes, Geniesse,
  Pappu, Leswing, and Pande]{wu2018moleculenet}
Wu,~Z.; Ramsundar,~B.; Feinberg,~E.; Gomes,~J.; Geniesse,~C.; Pappu,~A.~S.;
  Leswing,~K.; Pande,~V. MoleculeNet: a benchmark for molecular machine
  learning. \emph{Chem. Sci.} \textbf{2018}, \emph{9}, 513--530\relax
\mciteBstWouldAddEndPuncttrue
\mciteSetBstMidEndSepPunct{\mcitedefaultmidpunct}
{\mcitedefaultendpunct}{\mcitedefaultseppunct}\relax
\EndOfBibitem
\bibitem[Mayr \latin{et~al.}(2018)Mayr, Klambauer, Unterthiner, Steijaert,
  Wegner, Ceulemans, Clevert, and Hochreiter]{mayr2018large}
Mayr,~A.; Klambauer,~G.; Unterthiner,~T.; Steijaert,~M.; Wegner,~J.~K.;
  Ceulemans,~H.; Clevert,~D.-A.; Hochreiter,~S. Large-scale comparison of
  machine learning methods for drug target prediction on ChEMBL. \emph{Chemical
  science} \textbf{2018}, \emph{9}, 5441--5451\relax
\mciteBstWouldAddEndPuncttrue
\mciteSetBstMidEndSepPunct{\mcitedefaultmidpunct}
{\mcitedefaultendpunct}{\mcitedefaultseppunct}\relax
\EndOfBibitem
\bibitem[Duvenaud \latin{et~al.}(2015)Duvenaud, Maclaurin,
  Aguilera-Iparraguirre, G\'{o}mez-Bombarelli, Hirzel, Aspuru-Guzik, and
  Adams]{duvenaud2015convolutional}
Duvenaud,~D.; Maclaurin,~D.; Aguilera-Iparraguirre,~J.;
  G\'{o}mez-Bombarelli,~R.; Hirzel,~T.; Aspuru-Guzik,~A.; Adams,~R.~P.
  Convolutional Networks on Graphs for Learning Molecular Fingerprints.
  Proceedings of the 28th International Conference on Neural Information
  Processing Systems - Volume 2. 2015; pp 2224--2232\relax
\mciteBstWouldAddEndPuncttrue
\mciteSetBstMidEndSepPunct{\mcitedefaultmidpunct}
{\mcitedefaultendpunct}{\mcitedefaultseppunct}\relax
\EndOfBibitem
\bibitem[Gal(2016)]{gal2016uncertainty}
Gal,~Y. Uncertainty in deep learning. Ph.D.\ thesis, University of Cambridge,
  2016\relax
\mciteBstWouldAddEndPuncttrue
\mciteSetBstMidEndSepPunct{\mcitedefaultmidpunct}
{\mcitedefaultendpunct}{\mcitedefaultseppunct}\relax
\EndOfBibitem
\bibitem[Kendall and Gal(2017)Kendall, and Gal]{Kendall2017}
Kendall,~A.; Gal,~Y. What Uncertainties Do We Need in Bayesian Deep Learning
  for Computer Vision? Proceedings of the 31st International Conference on
  Neural Information Processing Systems. 2017; pp 5580--5590\relax
\mciteBstWouldAddEndPuncttrue
\mciteSetBstMidEndSepPunct{\mcitedefaultmidpunct}
{\mcitedefaultendpunct}{\mcitedefaultseppunct}\relax
\EndOfBibitem
\bibitem[Zhang and Lee(2019)Zhang, and Lee]{Zhang2019}
Zhang,~Y.; Lee,~A.~A. Bayesian semi-supervised learning for
  uncertainty-calibrated prediction of molecular properties and active
  learning. \emph{Chem. Sci.} \textbf{2019}, \emph{10}, 8154--8163\relax
\mciteBstWouldAddEndPuncttrue
\mciteSetBstMidEndSepPunct{\mcitedefaultmidpunct}
{\mcitedefaultendpunct}{\mcitedefaultseppunct}\relax
\EndOfBibitem
\bibitem[Proppe and Reiher(2017)Proppe, and Reiher]{proppe2017reliable}
Proppe,~J.; Reiher,~M. Reliable Estimation of Prediction Uncertainty for
  Physicochemical Property Models. \emph{Journal of chemical theory and
  computation} \textbf{2017}, \emph{13}, 3297--3317\relax
\mciteBstWouldAddEndPuncttrue
\mciteSetBstMidEndSepPunct{\mcitedefaultmidpunct}
{\mcitedefaultendpunct}{\mcitedefaultseppunct}\relax
\EndOfBibitem
\bibitem[Neal(1996)]{neal1995bayesian}
Neal,~R.~M. \emph{Bayesian Learning for Neural Networks}; Springer-Verlag,
  1996\relax
\mciteBstWouldAddEndPuncttrue
\mciteSetBstMidEndSepPunct{\mcitedefaultmidpunct}
{\mcitedefaultendpunct}{\mcitedefaultseppunct}\relax
\EndOfBibitem
\bibitem[Gal and Ghahramani(2016)Gal, and Ghahramani]{gal2016dropout}
Gal,~Y.; Ghahramani,~Z. Dropout As a Bayesian Approximation: Representing Model
  Uncertainty in Deep Learning. Proceedings of the 33rd International
  Conference on International Conference on Machine Learning - Volume 48. 2016;
  pp 1050--1059\relax
\mciteBstWouldAddEndPuncttrue
\mciteSetBstMidEndSepPunct{\mcitedefaultmidpunct}
{\mcitedefaultendpunct}{\mcitedefaultseppunct}\relax
\EndOfBibitem
\bibitem[Ryu \latin{et~al.}(2018)Ryu, Kwon, and Kim]{Ryu2018}
Ryu,~S.; Kwon,~Y.; Kim,~W.~Y. Uncertainty quantification of molecular property
  prediction using Bayesian neural network models. \emph{arXiv preprint
  arXiv:1905.06945} \textbf{2018}, \relax
\mciteBstWouldAddEndPunctfalse
\mciteSetBstMidEndSepPunct{\mcitedefaultmidpunct}
{}{\mcitedefaultseppunct}\relax
\EndOfBibitem
\bibitem[Lakshminarayanan \latin{et~al.}(2017)Lakshminarayanan, Pritzel, and
  Blundell]{lakshminarayanan2017simple}
Lakshminarayanan,~B.; Pritzel,~A.; Blundell,~C. Simple and Scalable Predictive
  Uncertainty Estimation Using Deep Ensembles. Proceedings of the 31st
  International Conference on Neural Information Processing Systems. 2017; pp
  6405--6416\relax
\mciteBstWouldAddEndPuncttrue
\mciteSetBstMidEndSepPunct{\mcitedefaultmidpunct}
{\mcitedefaultendpunct}{\mcitedefaultseppunct}\relax
\EndOfBibitem
\bibitem[Fauw \latin{et~al.}(2018)Fauw, Ledsam, Romera-Paredes, Nikolov,
  Tomasev, Blackwell, Askham, Glorot, O’Donoghue, Visentin, van~den
  Driessche, Lakshminarayanan, Meyer, Mackinder, Bouton, Ayoub, Chopra, King,
  Karthikesalingam, Hughes, Raine, Hughes, Sim, Egan, Tufail, Montgomery,
  Hassabis, Rees, Back, Khaw, Suleyman, Cornebise, Keane, and
  Ronneberger]{Ron18}
Fauw,~J.~D. \latin{et~al.}  Clinically applicable deep learning for diagnosis
  and referral in retinal disease. \emph{Nature Medicine} \textbf{2018},
  \emph{24}, 1342--1350\relax
\mciteBstWouldAddEndPuncttrue
\mciteSetBstMidEndSepPunct{\mcitedefaultmidpunct}
{\mcitedefaultendpunct}{\mcitedefaultseppunct}\relax
\EndOfBibitem
\bibitem[Tomašev \latin{et~al.}(2019)Tomašev, Glorot, Rae, Zielinski, Askham,
  Saraiva, Mottram, Meyer, Ravuri, Protsyuk, Connell, Hughes, Karthikesalingam,
  Cornebise, Montgomery, Rees, Laing, Baker, Peterson, Reeves, Hassabis, King,
  Suleyman, Back, Nielson, Ledsam, and Mohamed]{Tomasev2019116}
Tomašev,~N. \latin{et~al.}  A clinically applicable approach to continuous
  prediction of future acute kidney injury. \emph{Nature} \textbf{2019},
  \emph{572}, 116--119\relax
\mciteBstWouldAddEndPuncttrue
\mciteSetBstMidEndSepPunct{\mcitedefaultmidpunct}
{\mcitedefaultendpunct}{\mcitedefaultseppunct}\relax
\EndOfBibitem
\bibitem[Duvenaud \latin{et~al.}(2016)Duvenaud, Maclaurin, and
  Adams]{duvenaud2016early}
Duvenaud,~D.; Maclaurin,~D.; Adams,~R.~P. Early Stopping as Nonparametric
  Variational Inference. Proceedings of the 19th International Conference on
  Artificial Intelligence and Statistics, {AISTATS} 2016. 2016; pp
  1070--1077\relax
\mciteBstWouldAddEndPuncttrue
\mciteSetBstMidEndSepPunct{\mcitedefaultmidpunct}
{\mcitedefaultendpunct}{\mcitedefaultseppunct}\relax
\EndOfBibitem
\bibitem[Pearce \latin{et~al.}(2018)Pearce, Zaki, Brintrup, and
  Neel]{pearce2018uncertainty}
Pearce,~T.; Zaki,~M.; Brintrup,~A.; Neel,~A. Uncertainty in neural networks:
  Bayesian ensembling. \emph{arXiv preprint arXiv:1810.05546} \textbf{2018},
  \relax
\mciteBstWouldAddEndPunctfalse
\mciteSetBstMidEndSepPunct{\mcitedefaultmidpunct}
{}{\mcitedefaultseppunct}\relax
\EndOfBibitem
\bibitem[Gustafsson \latin{et~al.}(2019)Gustafsson, Danelljan, and
  Sch{\"o}n]{gustafsson2019evaluating}
Gustafsson,~F.~K.; Danelljan,~M.; Sch{\"o}n,~T.~B. Evaluating Scalable Bayesian
  Deep Learning Methods for Robust Computer Vision. \emph{arXiv preprint
  arXiv:1906.01620} \textbf{2019}, \relax
\mciteBstWouldAddEndPunctfalse
\mciteSetBstMidEndSepPunct{\mcitedefaultmidpunct}
{}{\mcitedefaultseppunct}\relax
\EndOfBibitem
\bibitem[Guo \latin{et~al.}(2017)Guo, Pleiss, Sun, and
  Weinberger]{guo2017calibration}
Guo,~C.; Pleiss,~G.; Sun,~Y.; Weinberger,~K.~Q. On Calibration of Modern Neural
  Networks. Proceedings of the 34th International Conference on Machine
  Learning - Volume 70. 2017; pp 1321--1330\relax
\mciteBstWouldAddEndPuncttrue
\mciteSetBstMidEndSepPunct{\mcitedefaultmidpunct}
{\mcitedefaultendpunct}{\mcitedefaultseppunct}\relax
\EndOfBibitem
\bibitem[Ilg \latin{et~al.}(2018)Ilg, Cicek, Galesso, Klein, Makansi, Hutter,
  and Brox]{ilg2018uncertainty}
Ilg,~E.; Cicek,~O.; Galesso,~S.; Klein,~A.; Makansi,~O.; Hutter,~F.; Brox,~T.
  Uncertainty estimates and multi-hypotheses networks for optical flow.
  Proceedings of the European Conference on Computer Vision (ECCV). 2018; pp
  652--667\relax
\mciteBstWouldAddEndPuncttrue
\mciteSetBstMidEndSepPunct{\mcitedefaultmidpunct}
{\mcitedefaultendpunct}{\mcitedefaultseppunct}\relax
\EndOfBibitem
\bibitem[Mukhoti and Gal(2018)Mukhoti, and Gal]{mukhoti2018evaluating}
Mukhoti,~J.; Gal,~Y. Evaluating Bayesian Deep Learning Methods for Semantic
  Segmentation. \emph{arXiv preprint arXiv:1811.12709} \textbf{2018}, \relax
\mciteBstWouldAddEndPunctfalse
\mciteSetBstMidEndSepPunct{\mcitedefaultmidpunct}
{}{\mcitedefaultseppunct}\relax
\EndOfBibitem
\bibitem[Kuleshov \latin{et~al.}(2018)Kuleshov, Fenner, and
  Ermon]{kuleshov2018accurate}
Kuleshov,~V.; Fenner,~N.; Ermon,~S. Accurate Uncertainties for Deep Learning
  Using Calibrated Regression. Proceedings of the 35th International Conference
  on Machine Learning. 2018; pp 2796--2804\relax
\mciteBstWouldAddEndPuncttrue
\mciteSetBstMidEndSepPunct{\mcitedefaultmidpunct}
{\mcitedefaultendpunct}{\mcitedefaultseppunct}\relax
\EndOfBibitem
\bibitem[Levi \latin{et~al.}(2019)Levi, Gispan, Giladi, and
  Fetaya]{levi2019evaluating}
Levi,~D.; Gispan,~L.; Giladi,~N.; Fetaya,~E. Evaluating and Calibrating
  Uncertainty Prediction in Regression Tasks. \emph{arXiv preprint
  arXiv:1905.11659} \textbf{2019}, \relax
\mciteBstWouldAddEndPunctfalse
\mciteSetBstMidEndSepPunct{\mcitedefaultmidpunct}
{}{\mcitedefaultseppunct}\relax
\EndOfBibitem
\bibitem[Coley \latin{et~al.}(2017)Coley, Barzilay, Green, Jaakkola, and
  Jensen]{coley2017convolutional}
Coley,~C.~W.; Barzilay,~R.; Green,~W.~H.; Jaakkola,~T.~S.; Jensen,~K.~F.
  Convolutional embedding of attributed molecular graphs for physical property
  prediction. \emph{Journal of chemical information and modeling}
  \textbf{2017}, \emph{57}, 1757--1772\relax
\mciteBstWouldAddEndPuncttrue
\mciteSetBstMidEndSepPunct{\mcitedefaultmidpunct}
{\mcitedefaultendpunct}{\mcitedefaultseppunct}\relax
\EndOfBibitem
\bibitem[Kendall \latin{et~al.}(2018)Kendall, Gal, and
  Cipolla]{kendall2018multi}
Kendall,~A.; Gal,~Y.; Cipolla,~R. Multi-task learning using uncertainty to
  weigh losses for scene geometry and semantics. Proceedings of the IEEE
  Conference on Computer Vision and Pattern Recognition. 2018; pp
  7482--7491\relax
\mciteBstWouldAddEndPuncttrue
\mciteSetBstMidEndSepPunct{\mcitedefaultmidpunct}
{\mcitedefaultendpunct}{\mcitedefaultseppunct}\relax
\EndOfBibitem
\bibitem[Le \latin{et~al.}(2005)Le, Smola, and Canu]{le2005heteroscedastic}
Le,~Q.~V.; Smola,~A.~J.; Canu,~S. Heteroscedastic Gaussian process regression.
  Proceedings of the 22nd international conference on Machine learning. 2005;
  pp 489--496\relax
\mciteBstWouldAddEndPuncttrue
\mciteSetBstMidEndSepPunct{\mcitedefaultmidpunct}
{\mcitedefaultendpunct}{\mcitedefaultseppunct}\relax
\EndOfBibitem
\bibitem[Nix and Weigend(1994)Nix, and Weigend]{nix1994estimating}
Nix,~D.~A.; Weigend,~A.~S. Estimating the mean and variance of the target
  probability distribution. Proceedings of 1994 IEEE International Conference
  on Neural Networks (ICNN'94). 1994; pp 55--60\relax
\mciteBstWouldAddEndPuncttrue
\mciteSetBstMidEndSepPunct{\mcitedefaultmidpunct}
{\mcitedefaultendpunct}{\mcitedefaultseppunct}\relax
\EndOfBibitem
\bibitem[Bishop(1994)]{bishop1994mixture}
Bishop,~C.~M. Mixture density networks. \textbf{1994}, \relax
\mciteBstWouldAddEndPunctfalse
\mciteSetBstMidEndSepPunct{\mcitedefaultmidpunct}
{}{\mcitedefaultseppunct}\relax
\EndOfBibitem
\bibitem[Choi \latin{et~al.}(2018)Choi, Lee, Lim, and Oh]{choi2018uncertainty}
Choi,~S.; Lee,~K.; Lim,~S.; Oh,~S. Uncertainty-aware learning from
  demonstration using mixture density networks with sampling-free variance
  modeling. 2018 IEEE International Conference on Robotics and Automation
  (ICRA). 2018; pp 6915--6922\relax
\mciteBstWouldAddEndPuncttrue
\mciteSetBstMidEndSepPunct{\mcitedefaultmidpunct}
{\mcitedefaultendpunct}{\mcitedefaultseppunct}\relax
\EndOfBibitem
\bibitem[Kristiadi and Fischer(2019)Kristiadi, and
  Fischer]{kristiadi2019predictive}
Kristiadi,~A.; Fischer,~A. Predictive Uncertainty Quantification with Compound
  Density Networks. \emph{arXiv preprint arXiv:1902.01080} \textbf{2019},
  \relax
\mciteBstWouldAddEndPunctfalse
\mciteSetBstMidEndSepPunct{\mcitedefaultmidpunct}
{}{\mcitedefaultseppunct}\relax
\EndOfBibitem
\bibitem[Ma \latin{et~al.}(2015)Ma, Chen, and Fox]{NIPS2015_5891}
Ma,~Y.-A.; Chen,~T.; Fox,~E.~B. A Complete Recipe for Stochastic Gradient MCMC.
  Proceedings of the 28th International Conference on Neural Information
  Processing Systems - Volume 2. 2015; pp 2917--2925\relax
\mciteBstWouldAddEndPuncttrue
\mciteSetBstMidEndSepPunct{\mcitedefaultmidpunct}
{\mcitedefaultendpunct}{\mcitedefaultseppunct}\relax
\EndOfBibitem
\bibitem[Zhang \latin{et~al.}(2019)Zhang, Li, Zhang, Chen, and
  Wilson]{zhang2019cyclical}
Zhang,~R.; Li,~C.; Zhang,~J.; Chen,~C.; Wilson,~A.~G. Cyclical stochastic
  gradient MCMC for Bayesian deep learning. \emph{arXiv preprint
  arXiv:1902.03932} \textbf{2019}, \relax
\mciteBstWouldAddEndPunctfalse
\mciteSetBstMidEndSepPunct{\mcitedefaultmidpunct}
{}{\mcitedefaultseppunct}\relax
\EndOfBibitem
\bibitem[Graves(2011)]{graves2011practical}
Graves,~A. Practical Variational Inference for Neural Networks. Proceedings of
  the 24th International Conference on Neural Information Processing Systems.
  2011; pp 2348--2356\relax
\mciteBstWouldAddEndPuncttrue
\mciteSetBstMidEndSepPunct{\mcitedefaultmidpunct}
{\mcitedefaultendpunct}{\mcitedefaultseppunct}\relax
\EndOfBibitem
\bibitem[Hern\'{a}ndez-Lobato and Adams(2015)Hern\'{a}ndez-Lobato, and
  Adams]{hernandez2015probabilistic}
Hern\'{a}ndez-Lobato,~J.~M.; Adams,~R.~P. Probabilistic Backpropagation for
  Scalable Learning of Bayesian Neural Networks. Proceedings of the 32Nd
  International Conference on International Conference on Machine Learning -
  Volume 37. 2015; pp 1861--1869\relax
\mciteBstWouldAddEndPuncttrue
\mciteSetBstMidEndSepPunct{\mcitedefaultmidpunct}
{\mcitedefaultendpunct}{\mcitedefaultseppunct}\relax
\EndOfBibitem
\bibitem[Liu and Wang(2016)Liu, and Wang]{liu2016stein}
Liu,~Q.; Wang,~D. Stein Variational Gradient Descent: A General Purpose
  Bayesian Inference Algorithm. Proceedings of the 30th International
  Conference on Neural Information Processing Systems. 2016; pp
  2378--2386\relax
\mciteBstWouldAddEndPuncttrue
\mciteSetBstMidEndSepPunct{\mcitedefaultmidpunct}
{\mcitedefaultendpunct}{\mcitedefaultseppunct}\relax
\EndOfBibitem
\bibitem[Li \latin{et~al.}(2019)Li, Han, Grambow, and Green]{Li2019}
Li,~Y.-P.; Han,~K.; Grambow,~C.~A.; Green,~W.~H. Self-Evolving Machine: A
  Continuously Improving Model for Molecular Thermochemistry. \emph{The Journal
  of Physical Chemistry A} \textbf{2019}, \emph{123}, 2142--2152\relax
\mciteBstWouldAddEndPuncttrue
\mciteSetBstMidEndSepPunct{\mcitedefaultmidpunct}
{\mcitedefaultendpunct}{\mcitedefaultseppunct}\relax
\EndOfBibitem
\bibitem[Smith \latin{et~al.}(2018)Smith, Nebgen, Lubbers, Isayev, and
  Roitberg]{smith2018less}
Smith,~J.~S.; Nebgen,~B.; Lubbers,~N.; Isayev,~O.; Roitberg,~A.~E. Less is
  more: Sampling chemical space with active learning. \emph{The Journal of
  chemical physics} \textbf{2018}, \emph{148}, 241733\relax
\mciteBstWouldAddEndPuncttrue
\mciteSetBstMidEndSepPunct{\mcitedefaultmidpunct}
{\mcitedefaultendpunct}{\mcitedefaultseppunct}\relax
\EndOfBibitem
\bibitem[Peterson \latin{et~al.}(2017)Peterson, Christensen, and
  Khorshidi]{peterson2017addressing}
Peterson,~A.~A.; Christensen,~R.; Khorshidi,~A. Addressing uncertainty in
  atomistic machine learning. \emph{Physical Chemistry Chemical Physics}
  \textbf{2017}, \emph{19}, 10978--10985\relax
\mciteBstWouldAddEndPuncttrue
\mciteSetBstMidEndSepPunct{\mcitedefaultmidpunct}
{\mcitedefaultendpunct}{\mcitedefaultseppunct}\relax
\EndOfBibitem
\bibitem[Gal \latin{et~al.}(2017)Gal, Hron, and Kendall]{gal2017concrete}
Gal,~Y.; Hron,~J.; Kendall,~A. Concrete dropout. Proceedings of the 31st
  International Conference on Neural Information Processing Systems. 2017; pp
  3584--3593\relax
\mciteBstWouldAddEndPuncttrue
\mciteSetBstMidEndSepPunct{\mcitedefaultmidpunct}
{\mcitedefaultendpunct}{\mcitedefaultseppunct}\relax
\EndOfBibitem
\bibitem[Srivastava \latin{et~al.}(2014)Srivastava, Hinton, Krizhevsky,
  Sutskever, and Salakhutdinov]{srivastava2014dropout}
Srivastava,~N.; Hinton,~G.; Krizhevsky,~A.; Sutskever,~I.; Salakhutdinov,~R.
  Dropout: a simple way to prevent neural networks from overfitting. \emph{The
  Journal of Machine Learning Research} \textbf{2014}, \emph{15},
  1929--1958\relax
\mciteBstWouldAddEndPuncttrue
\mciteSetBstMidEndSepPunct{\mcitedefaultmidpunct}
{\mcitedefaultendpunct}{\mcitedefaultseppunct}\relax
\EndOfBibitem
\bibitem[Dietterich(2000)]{dietterich2000ensemble}
Dietterich,~T.~G. Ensemble Methods in Machine Learning. Proceedings of the
  First International Workshop on Multiple Classifier Systems. 2000; pp
  1--15\relax
\mciteBstWouldAddEndPuncttrue
\mciteSetBstMidEndSepPunct{\mcitedefaultmidpunct}
{\mcitedefaultendpunct}{\mcitedefaultseppunct}\relax
\EndOfBibitem
\bibitem[Goodfellow \latin{et~al.}(2016)Goodfellow, Bengio, and
  Courville]{Goodfellow2016}
Goodfellow,~I.; Bengio,~Y.; Courville,~A. \emph{{Deep Learning}}; MIT Press,
  2016\relax
\mciteBstWouldAddEndPuncttrue
\mciteSetBstMidEndSepPunct{\mcitedefaultmidpunct}
{\mcitedefaultendpunct}{\mcitedefaultseppunct}\relax
\EndOfBibitem
\bibitem[Mandt \latin{et~al.}(2017)Mandt, Hoffman, and
  Blei]{mandt2017stochastic}
Mandt,~S.; Hoffman,~M.~D.; Blei,~D.~M. Stochastic gradient descent as
  approximate bayesian inference. \emph{The Journal of Machine Learning
  Research} \textbf{2017}, \emph{18}, 4873--4907\relax
\mciteBstWouldAddEndPuncttrue
\mciteSetBstMidEndSepPunct{\mcitedefaultmidpunct}
{\mcitedefaultendpunct}{\mcitedefaultseppunct}\relax
\EndOfBibitem
\bibitem[Niculescu-Mizil and Caruana(2005)Niculescu-Mizil, and
  Caruana]{niculescu2005predicting}
Niculescu-Mizil,~A.; Caruana,~R. Predicting good probabilities with supervised
  learning. Proceedings of the 22nd international conference on Machine
  learning. 2005; pp 625--632\relax
\mciteBstWouldAddEndPuncttrue
\mciteSetBstMidEndSepPunct{\mcitedefaultmidpunct}
{\mcitedefaultendpunct}{\mcitedefaultseppunct}\relax
\EndOfBibitem
\bibitem[DeGroot and Fienberg(1983)DeGroot, and
  Fienberg]{degroot1983comparison}
DeGroot,~M.~H.; Fienberg,~S.~E. The comparison and evaluation of forecasters.
  \emph{Journal of the Royal Statistical Society: Series D (The Statistician)}
  \textbf{1983}, \emph{32}, 12--22\relax
\mciteBstWouldAddEndPuncttrue
\mciteSetBstMidEndSepPunct{\mcitedefaultmidpunct}
{\mcitedefaultendpunct}{\mcitedefaultseppunct}\relax
\EndOfBibitem
\bibitem[Naeini \latin{et~al.}(2015)Naeini, Cooper, and
  Hauskrecht]{naeini2015obtaining}
Naeini,~M.~P.; Cooper,~G.~F.; Hauskrecht,~M. Obtaining Well Calibrated
  Probabilities Using Bayesian Binning. Proceedings of the Twenty-Ninth AAAI
  Conference on Artificial Intelligence. 2015; pp 2901--2907\relax
\mciteBstWouldAddEndPuncttrue
\mciteSetBstMidEndSepPunct{\mcitedefaultmidpunct}
{\mcitedefaultendpunct}{\mcitedefaultseppunct}\relax
\EndOfBibitem
\bibitem[Gneiting \latin{et~al.}(2007)Gneiting, Balabdaoui, and
  Raftery]{gneiting2007probabilistic}
Gneiting,~T.; Balabdaoui,~F.; Raftery,~A.~E. Probabilistic forecasts,
  calibration and sharpness. \emph{Journal of the Royal Statistical Society:
  Series B (Statistical Methodology)} \textbf{2007}, \emph{69}, 243--268\relax
\mciteBstWouldAddEndPuncttrue
\mciteSetBstMidEndSepPunct{\mcitedefaultmidpunct}
{\mcitedefaultendpunct}{\mcitedefaultseppunct}\relax
\EndOfBibitem
\bibitem[Beluch \latin{et~al.}(2018)Beluch, Genewein, Nurnberger, and
  Kohler]{beluch2018power}
Beluch,~W.~H.; Genewein,~T.; Nurnberger,~A.; Kohler,~J.~M. The Power of
  Ensembles for Active Learning in Image Classification. 2018 IEEE/CVF
  Conference on Computer Vision and Pattern Recognition. 2018; pp
  9368--9377\relax
\mciteBstWouldAddEndPuncttrue
\mciteSetBstMidEndSepPunct{\mcitedefaultmidpunct}
{\mcitedefaultendpunct}{\mcitedefaultseppunct}\relax
\EndOfBibitem
\bibitem[Ramakrishnan \latin{et~al.}(2014)Ramakrishnan, Dral, Rupp, and von
  Lilienfeld]{Ramakrishnan2014}
Ramakrishnan,~R.; Dral,~P.~O.; Rupp,~M.; von Lilienfeld,~O.~A. {Quantum
  chemistry structures and properties of 134 kilo molecules}. \emph{Scientific
  Data} \textbf{2014}, \emph{1}, 140022\relax
\mciteBstWouldAddEndPuncttrue
\mciteSetBstMidEndSepPunct{\mcitedefaultmidpunct}
{\mcitedefaultendpunct}{\mcitedefaultseppunct}\relax
\EndOfBibitem
\bibitem[Cohen \latin{et~al.}(2012)Cohen, Mori-Sánchez, and
  Yang]{cohen_challenges_2012}
Cohen,~A.~J.; Mori-Sánchez,~P.; Yang,~W. Challenges for {Density} {Functional}
  {Theory}. \emph{Chemical Reviews} \textbf{2012}, \emph{112}, 289--320\relax
\mciteBstWouldAddEndPuncttrue
\mciteSetBstMidEndSepPunct{\mcitedefaultmidpunct}
{\mcitedefaultendpunct}{\mcitedefaultseppunct}\relax
\EndOfBibitem
\bibitem[Simm and Reiher(2016)Simm, and Reiher]{simm_systematic_2016}
Simm,~G.~N.; Reiher,~M. Systematic {Error} {Estimation} for {Chemical}
  {Reaction} {Energies}. \emph{Journal of Chemical Theory and Computation}
  \textbf{2016}, \emph{12}, 2762--2773\relax
\mciteBstWouldAddEndPuncttrue
\mciteSetBstMidEndSepPunct{\mcitedefaultmidpunct}
{\mcitedefaultendpunct}{\mcitedefaultseppunct}\relax
\EndOfBibitem
\bibitem[Proppe \latin{et~al.}(2017)Proppe, Husch, Simm, and
  Reiher]{proppe_uncertainty_2017}
Proppe,~J.; Husch,~T.; Simm,~G.~N.; Reiher,~M. Uncertainty quantification for
  quantum chemical models of complex reaction networks. \emph{Faraday
  Discussions} \textbf{2017}, \emph{195}, 497--520\relax
\mciteBstWouldAddEndPuncttrue
\mciteSetBstMidEndSepPunct{\mcitedefaultmidpunct}
{\mcitedefaultendpunct}{\mcitedefaultseppunct}\relax
\EndOfBibitem
\bibitem[Li \latin{et~al.}(2016)Li, Bell, and
  Head-Gordon]{li_thermodynamics_2016}
Li,~Y.-P.; Bell,~A.~T.; Head-Gordon,~M. Thermodynamics of {Anharmonic}
  {Systems}: {Uncoupled} {Mode} {Approximations} for {Molecules}. \emph{Journal
  of Chemical Theory and Computation} \textbf{2016}, \emph{12},
  2861--2870\relax
\mciteBstWouldAddEndPuncttrue
\mciteSetBstMidEndSepPunct{\mcitedefaultmidpunct}
{\mcitedefaultendpunct}{\mcitedefaultseppunct}\relax
\EndOfBibitem
\bibitem[Grambow \latin{et~al.}(2019)Grambow, Li, and
  Green]{grambow_accurate_2019}
Grambow,~C.~A.; Li,~Y.-P.; Green,~W.~H. Accurate {Thermochemistry} with {Small}
  {Data} {Sets}: {A} {Bond} {Additivity} {Correction} and {Transfer} {Learning}
  {Approach}. \emph{The Journal of Physical Chemistry. A} \textbf{2019},
  \emph{123}, 5826--5835\relax
\mciteBstWouldAddEndPuncttrue
\mciteSetBstMidEndSepPunct{\mcitedefaultmidpunct}
{\mcitedefaultendpunct}{\mcitedefaultseppunct}\relax
\EndOfBibitem
\bibitem[Lee \latin{et~al.}(2015)Lee, Purushwalkam, Cogswell, Crandall, and
  Batra]{lee2015m}
Lee,~S.; Purushwalkam,~S.; Cogswell,~M.; Crandall,~D.; Batra,~D. Why M heads
  are better than one: Training a diverse ensemble of deep networks.
  \emph{arXiv preprint arXiv:1511.06314} \textbf{2015}, \relax
\mciteBstWouldAddEndPunctfalse
\mciteSetBstMidEndSepPunct{\mcitedefaultmidpunct}
{}{\mcitedefaultseppunct}\relax
\EndOfBibitem
\bibitem[Pearce \latin{et~al.}(2018)Pearce, Anastassacos, Zaki, and
  Neely]{pearce2018bayesian}
Pearce,~T.; Anastassacos,~N.; Zaki,~M.; Neely,~A. Bayesian Inference with
  Anchored Ensembles of Neural Networks, and Application to Reinforcement
  Learning. \emph{arXiv preprint arXiv:1805.11324} \textbf{2018}, \relax
\mciteBstWouldAddEndPunctfalse
\mciteSetBstMidEndSepPunct{\mcitedefaultmidpunct}
{}{\mcitedefaultseppunct}\relax
\EndOfBibitem
\end{mcitethebibliography}

\end{document}